\documentclass[runningheads]{llncs}
\usepackage{breakcites}

\usepackage{appendix}
\usepackage{array}
\usepackage{comment}
\usepackage{graphicx}
\usepackage{amsmath}
\usepackage{amssymb}
\usepackage[colorlinks]{hyperref}
\newcommand{\PAR}[1]{\vskip4pt \noindent{\bf #1~}}

\usepackage[colorlinks]{hyperref}
\usepackage{array}
\newcolumntype{P}[1]{>{\centering\arraybackslash}p{#1}}
\usepackage{tablefootnote}

\begin{document}

\pagestyle{headings}
\mainmatter
\def\ECCVSubNumber{6194}  

\title{Objects Can Move: 3D Change Detection by Geometric Transformation Consistency}


\titlerunning{Objects Can Move}
%
\author{Aikaterini Adam\inst{1,2} \and
Torsten Sattler\inst{1} \and
Konstantinos Karantzalos\inst{2}\and
Tomas Pajdla\inst{1}}
\authorrunning{A. Adam et al.}
%
\institute{Czech Institute of Informatics, Robotics and Cybernetics, CTU in Prague 
\email{\{Aikaterini.Adam,Torsten.Sattler,pajdla\}@cvut.cz} \and
National Technical University of Athens, Greece\\
\email{karank@scentral.ntua.gr}}
\maketitle

\begin{abstract}
AR/VR applications and robots need to know when the scene has changed. An example is when objects are moved, added, or removed from the scene. We propose a 3D object discovery method that is based only on scene changes. Our method does not need to encode any assumptions about what is an object, but rather discovers objects by exploiting their coherent move. Changes are initially detected as differences in the depth maps and segmented as objects if they undergo rigid motions. A graph cut optimization propagates the changing labels to geometrically consistent regions. Experiments show that our method achieves state-of-the-art performance on the 3RScan dataset against competitive baselines. The source code of our method can be found at \href{URL}{https://github.com/katadam/ObjectsCanMove}.  
\end{abstract}

\section{Introduction}

The ability to detect and interact with objects is critical to AR/VR applications and for multiple robotics tasks, such as surveillance, robotic manipulation, and maintaining order. All these tasks are operated in the same setting. Thus, the robot, or the  AR/VR device stores a reference map and builds a new map upon each revisit.  However, in-between the revisits, certain objects may have changed. Checking for scene consistency and detecting changes on an object-level can thus lead to 3D object discovery, without the need of labeled data. 

Motivated by the above, we explore an object discovery approach, based on examining scene consistency on an object-level and without using annotated data. We are aiming at discovering entities (objects) that have changed when revisiting a place. We show that it is possible to detect 3D objects purely geometrically, without a predefined notion of objects. The underlying idea is that objects, unlike the static background of a scene, can be moved. This is an intuitive definition of ``objectness'' that does not need any annonated data. 

Segmenting dynamic objects in temporal observations is a long-standing challenge. There are two ways to apply this idea: (1) segment objects from the background by actively observing their motion, e.g., by reconstructing dynamic objects during SLAM~\cite{wongrigidfusion}, or (2) revisit the same scene after a (longer) period and detect potential objects as changes between two maps~\cite{halber2019rescan}. We follow the latter approach, i.e., we model the problem as a change detection task. 

Detecting potential scene changes based on direct data analytics is a task attracting much attention since affordable 3D scanning technology \cite{glocker2013real,wald2019rio,armeni20163d} makes such data widely available. However, a straightforward approach to detecting changes between two scans based on voxel occupancy or inconsistency maps~\cite{palazzolo2018fast} would often miss changes,  e.g., when an object rotates around an axis passing through the object or when it is ``slid along itself''. An alternative approach employs the comparison of visual features and relies on photoconsistency constraints~\cite{taneja2011image}. Yet, this approach does not perform well in our setting since there can be significant illumination changes between the two maps. 

\begin{figure*}[b!]
\centering
\includegraphics[width=\textwidth,trim={0cm 0cm 0cm 1cm}]{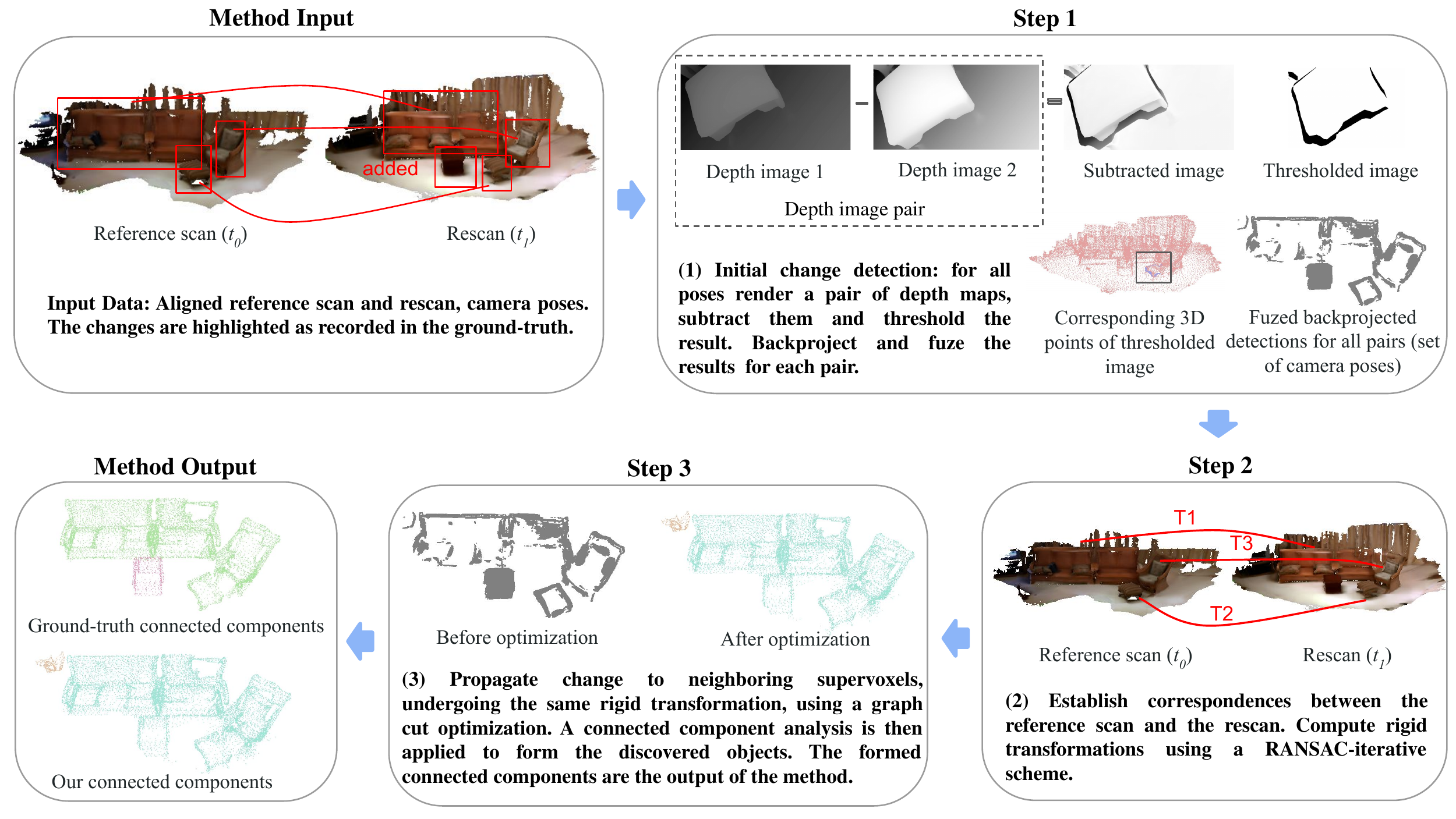} 
\caption{ Workflow of the proposed method:  given two scans recording changes and the associated camera poses, we discover all objects that have been added, moved, or removed from the scene. Initial geometric changes are detected as differences in depth maps (Step 1). The dominant transformations are then computed (Step 2). The initial set of detections is incomplete and thus refined, using a graph cut-based optimization on a supervoxel representation, propagating change to all regions undergoing the same transformation (Step 3). Discovered objects are presented as the extracted connected components of the refined detections}
\label{fig:concept}
\centering
\end{figure*}

To tackle the aforementioned shortcomings, we introduce a novel change detection framework, depicted in Figure \ref{fig:concept}, that uses geometric transformation consistency towards object discovery (i.e., change detection on an object-level). 3D objects are thus discovered without the need to encode what an object is.  
We consider a scenario where we have two 3D maps, i.e., a reference scan (recorded at time $t_{0}$) and a rescan (recorded at time $t_{1}$), of a scene, as well as the associated camera poses. Initial change detections are computed as differences in the depth maps. As shown in Figures~\ref{fig:concept} and~\ref{fig:overall}, the initial detected points mainly delineate the boundaries of the moved objects. To recover all parts, we propagate changes from regions where we can detect them to parts where no changes were seen, but which belong to the same object. Our local robust feature matching between parts of the two scans generates motion hypotheses for the scene parts, induced by the moved objects. These motions can measure consistency as scene parts that undergo the same rigid transformation.

\PAR{Contributions.} We introduce a novel 3D change detection framework via geometric transformation consistency. As change detection is performed on an object-level, this novel framework serves as an  object discovery method in 3D scenes, without needing any strong priors or definition of what objects are. We showcase that even though we target rigid objects/changes, our method can also handle non-rigid changes, as shown in Figure \ref{curtain}. The proposed method achieves state-of-the-art performance on the 3RScan dataset~\cite{wald2019rio}, against competitive baselines.

We evaluate our framework on the 3RScan dataset~\cite{wald2019rio}, initially designed for benchmarking object instance relocalization. Our evaluation shows the potential of the dataset to assess 3D change detection. We provide code to generate the ground truth annotations.

\section{Related Work} \label{related}
{\PAR{Change Detection.}} 3D Change detection is directly related to our method since the presented workflow is modeled in this concept.
Change detection has been traditionally treated mostly by geometric approaches \cite{taneja2013city,taneja2011image,palazzolo2017change,palazzolo2018fast,xiao2013change,ulusoy2014image}. Similar to our initial detection step, \cite{taneja2011image,taneja2013city,palazzolo2017change} detect changes based on inconsistency maps from RGB or depth projections.
Many change detection algorithms \cite{taneja2011image,palma2016detection} are based on the concept of initial change detection (e.g., though color consistency, comparing depth values, etc.), followed by propagating these detections to identify all regions that have  changed. \cite{taneja2011image,palma2016detection} propagate change using spatial and photoconsistency constraints. Our approach follows the same outline, but differs in the key step of change propagation, through a novel geometric twist. Thus, our method is illumination invariant and can be applied to complex, open-set environments under varying illumination conditions.

{\PAR{SLAM Methods for Dynamic Object Segmentation.}} When addressing dynamic scenes, tracking dynamic objects can be part of SLAM-based techniques. In \cite{ambrus2016unsupervised}, dynamic parts of the scenes are recovered and a classifier is trained on them to distinguish between static and non-static parts.  Semantic SLAM for dynamic environments is presented in \cite{brasch2018semantic,cui2019sof}. In \cite{runz2018maskfusion}, the authors first segment objects and track them separately. In a similar vein to our research, \cite{finman2013toward,herbst2011toward,mason2012object} aim at discovering objects through change observation on an object-level. However, these works build their methods upon a SLAM-based basis. Our method is complementary to SLAM-based techniques since these methods demand the recording of the object's actual movement in front of the camera. On the other hand, our method needs two 3D models (reference scan and rescan), and the associated camera poses, which are acquired over long time intervals. Thus, objects might have moved, appeared, or disappeared without their movement being explicitly recorded.

{\PAR{3D Object Discovery.}} Our problem can be conceived as a 3D object discovery technique when declaring as an object everything that can be moved, since movement is an inherent property of objects. Concerning unsupervised object discovery, the authors of \cite{karpathy2013object} focus on identifying  parts of the input mesh as candidate objects. They then  classify them as an object or clutter. More similarly to our work, \cite{langer2017fly,langer2020robust} extract as objects all the novel additions to the scene. Indeed, by scene comparison, they discover and label as an object anything that has been added between two scans. In contrast, our proposed method does not restrict itself only to added objects, but rather discovers all the objects that have changed (added, moved or removed).


\section{Detection via Geometric Consistency}
Our method aims at detecting changes on an object-level, thus leading to object discovery, without relying on annotated data. Given two 3D scans, i.e., a reference scan and a rescan, and the associated camera poses, we propose three discrete steps, as illustrated in Figure~\ref{fig:concept}: (1) initial change detection, i.e., compute the locations where a change might have occured, (2) compute dominant transformations, and (3) graph optimization to ensure geometrical transformation consistency. Differences in depth maps provide an initial but incomplete set of detections later refined using a graph cut-based optimization. The central insight is that scene parts that belong to the same object should undergo the same physical transformation, which we model through a novel geometric transformation consistency measure.  A connected component analysis is then applied to form the discovered objects. 

Initial changes are calculated by depth map comparison. Given a reference scan $S$ of the scene and a rescan $R$ aligned to each other, we render and subtract depth maps. Their subtraction records changing depth values and thus indicates changed regions. However, due to the way the objects move, it is difficult to retrieve the whole object via this single step (as illustrated in Figure \ref{fig:overall}). To tackle this limitation, we integrate graph optimization \cite{landrieu2017cut}, performed on supervoxels \cite{papon2013voxel}.  Instead of using a simple voxel representation of the scene, we firstly compute supervoxels, i.e., irregular clusters of 3D points sharing common geometrical and color characteristics. Optimizing this representation leads to more accurate results, since supervoxels separate the 3D space into elements, by clustering points with same properties. This is not the case for voxels that are created solely on spatial relations of the 3D points. Moreover, as supervoxels are irregular patches of 3D points, they can preserve objects' boundaries, contrary to the simple voxel representation.

Graph optimization aims to enforce consistency for all the regions undergoing the same rigid transformation. This will help us discover parts of the moving object that may have been missed during the initial detection step. The change is propagated to all the supervoxels undergoing the same movement. From the above, it is clear that two steps are needed before the optimization: (1) initial change detection, and (2) computation of all the dominant transformations induced by moved objects. Towards the latter goal, learned descriptors \cite{phan2018dgcnn} are extracted for each point in the scans. We use a pre-trained model, trained on a completely different task (i.e., semantic segmentation). Matches are then computed using nearest neighbor search \cite{JDH17}. The resulting correspondences are used to calculate the 3D transformations.

\PAR{Scan Alignment.} Works and datasets \cite{wald2019rio,halber2019rescan} exploring changing indoor scenes demand the two scans to be registered. These datasets provide information for the alignment since registering the scans is outside the scope of their research. Similarly to these works, we use the initial alignment provided by the dataset, which was obtained via manual annotations and is imperfect. In practical applications, the alignment could be provided by re-localization to the previous scan, or by estimating the overall transformation via feature matching \cite{bai2021pointdsc}.


\subsection{Initial Change Detection} \label{initial_dect}
The first step of our method is identifying changing regions, which we will refine via a graph optimization. Initial change detection is based on depth map comparison. We render depth maps $\mathcal{D}_{S,1,..,N}$, $\mathcal{D}_{R,1,..,N}$ from the reference scan $S$ and the rescan $R$ respectively, for all the viewpoints $i = 1,2,..,N$, using the $\mathbf{P}_{1,..,N}$ projection matrices. Multiple poses cover the whole 3D scene. We use the same poses to render both the reference scan $S$ and the rescan $R$, as we assume that both scans are already aligned, even if captured from different viewpoints.  We render the depth images rather than simply use raw depth measurements captured by a device to ensure the best possible quality. 
Moreover, we use depth maps instead of directly working on the mesh, allowing us to handle occlusions and partial observations more naturally than in the 3D space. Working on depth images  provides information about free space, which is not directly included in the mesh. Indeed, rendering depth allows us to know if the corresponding 3D region has been scanned or not. If one of the depth images does not contain information, we exclude this region from the initial change detection. This procedure is not straightforward in the mesh, as an intermediate step such as calculating the bounding box of the scan or computing overlapping regions would be needed to ensure that partial observations and free space is taken into consideration. Rendered maps from the reference scan and the rescan are shown in Figure \ref{fig:concept}.  Lighter regions correspond to regions that lie closer to the camera. The paired depth maps are subtracted, and the result is thresholded using \cite{barron2020generalization}. 

The result is a binary mask, encoding information about changing regions, which are back-projected  to the 3D space: 

\begin{equation} \label{eq:8}
[X,Y,Z]^{T} = \mathbf{R}^{T}\cdot( \mathbf{K}^{-1} \cdot [x,y,1]^{T}\cdot D(x,y)-\mathbf{Tr} )\enspace ,
\end{equation}
where $[X,Y,Z]$ stands for the world-coordinates of the 3D point, $\mathbf{R}$ for the rotation matrix, $\mathbf{K}$ for the calibration matrix and $\mathbf{Tr}$ for the translation vector, all forming the projection matrix $\mathbf{P} = \mathbf{K}[\mathbf{R}|\mathbf{Tr}]$. Vector $[x,y,1]$ represents the pixel coordinates and $D_{1,..,N}(x,y) $ the depth value stemming from the combination of the depth of the reference scan and the depth of the rescan: 
\begin{equation} \label{eq:9}
D(x,y)_{1,..,N} = \begin{cases}
D(x,y)_{S,1,..,N}   & \text{ if   $D(x,y)_{S,1,..,N}<D(x,y)_{R,1,..,N}$ } \\ 
D(x,y)_{R,1,..,N} & \text{ if   $D(x,y)_{S,1,..,N}>D(x,y)_{R,1,..,N}$ }
\end{cases}.
\end{equation}
$D(x,y)_{1,..,N}$ represents the depth value at position $(x,y)$ for the combined masks $1,..,N$, $D(x,y)_{S,1,..,N}$ the depth value at position $(x,y)$ for the $1,..,N$ rendered depth maps of the reference scan $S$ and $D(x,y)_{R,1,..,N}$ the depth value of the rendered $1,..,N$ depth maps of the rescan $R$. 
This formulation always selects the closest objects to the camera. After experiments, we concluded that in  97\% of examined cases, the smaller depth value corresponds to an object. In contrast, the larger depth corresponds to the static background. 
Figures \ref{fig:concept} and \ref{fig:overall} depict all the initial points labeled as changing for example scenes. 
Finally, as the graph optimization is applied to the supervoxel representation, the supervoxels for the scan and the rescan are extracted \cite{papon2013voxel} and the number of changing points belonging to each supervoxel is computed.

\subsection{Computing Dominant Transformations} \label{Transformations}
As seen in Figures~\ref{fig:concept}-Before optimization,~\ref{fig:overall}-Initial detected points, our initial detection step may miss changes
due to occluded parts of the objects, objects partially captured in one of the scans or due to the way objects have moved. Indeed, when an object is only slightly moved or rotated, there can be regions where the depth values do not change,
e.g., when a couch is only slightly shifted.  As a result, the initial detections might only cover part of the object. To this end, we use a graph optimization to propagate change detection to the rest of the objects based on consistency under geometric transformations $\mathbf{T}$.  We thus compute the different 3D rigid transformations $\mathbf{T}$, induced by the moving objects. Towards that, we match feature descriptors between scans. 

Descriptors can be computed using hand-crafted features, such as FPFH \cite{rusu2009fast} and SHOT \cite{tombari2010unique}, or learned features. In our case, we use features extracted from the encoder part of a pre-trained deep network. A forward-pass was deployed to densely extract descriptors from the scans, using the weights of pre-trained models on a semantic segmentation task. The models we are using are trained for a completely different task and dataset. Yet, using learned features does not affect our assertion of presenting an unsupervised method.

Correspondences were then computed over the entire scene, using the 
nearest neighbor search. Visualizing correspondences for the different features showed that the pre-trained Dynamic Graph CNN (DGCNN)~\cite{wang2019dynamic} had the best preliminary results. To remove outliers from the matches, and given that we want to establish correspondences only between moving objects, we eliminate correspondences lying within a predefined distance of each other in 3D, as these points are considered part of the static background. All the valid correspondences are then employed to compute the potential transformations using RANSAC~\cite{bolles1981ransac}. We iteratively apply RANSAC on the remaining set of matches after removing the inliers of the previous estimate and stop once less than three matches remain. Since this method will generate more transformations $\mathbf{T}$ than the real ones due to limitations in establishing correspondences, we will continue selecting the top $k$ transformations $\mathbf{T}$, with the most inliers, to propagate change during graph optimization. 

\subsection{Supervoxel Graph Optimization} \label{optimization}
From the initial change detection (Section \ref{initial_dect}), we obtain an initial soft labeling $L$, based on the fraction of changing 3D points belonging to each supervoxel. Supervoxels with more points labeled as changing, during the initial change detection, are more likely to belong to an object. Thus, the initial labeling $L$ determines the probability $\rho$ of each supervoxel $v_i$ to be labeled as changing $\rho(v_i, l_i = 1)$, or non changing  $\rho(v_i, l_i = 0)$ as:

\begin{align} \label{eq:soft labeling 1}
\rho(v_i, l_i = 1)  = \begin{cases}
0.8 & \text{if changing points $\in  {v_{i}}$} \\
0.5 & \text{if changing points $\notin  {v_{i}}$} \\
\end{cases}\enspace ,\\
\rho(v_i, l_i = 0)  =  \begin{cases}
0.2 & \text{if changing points $\in  {v_{i}}$} \\
0.5 & \text{if changing points $\notin  {v_{i}}$} \\
\end{cases}.
\label{eq:soft labeling 2}\enspace 
\end{align}

The weights used in Equations \ref{eq:soft labeling 1} and \ref{eq:soft labeling 2} were chosen based on experiments with different values on a set of scans used for tuning hyperparameters (cf. Section~\ref{Experimental}).  From the above, it is clear that we treat supervoxels with no detected changing points as equally likely of having changed. Indeed, supervoxels without any detected changing points do not necessarily correspond to static scene parts. We thus decide whether a supervoxel belongs to a moving object or not by solving a graph optimization problem~\cite{landrieu2017cut} that allows us to propagate changes between supervoxels, conditioned on a rigid transformation $\mathbf{T}$. 

To deploy the optimization, we create an undirected graph $\mathcal{G}=({V},{E})$. Each node $v\in{V}$ corresponds to a supervoxel in the scene. Two nodes $v_i$ and $v_j$ are connected through an edge $\{v_i, v_j\} \in {E}$ if the corresponding supervoxels are adjacent to each other. Given a rigid transformation $\mathbf{T}$ (cf. Section~\ref{Transformations}) between the rescan and the reference, our goal is to compute an optimised binary labeling $\mathcal{L^{*}} = \{l^{*}_i\}_i$. This labeling  indicates for each supervoxel whether  it belongs to a changing object consistent with $\mathbf{T}$ (and thus is labeled as $l^{*}_i = 1 $) or not ($l^{*}_i = 0$). 
We compute this labeling by solving the graph optimization problem~\cite{landrieu2017cut}:
\begin{equation} 
L^{*} \in {\underset{Q\in\Omega^{v}}{\arg\min}} \{\Phi(L,Q)+\lambda\Psi(Q)\} \enspace .
\label{eq:optimized labeling}
\end{equation} 

$\Phi$ stands for the fidelity term (here, we use the Kullback-Leibler fidelity function \cite{kullback1951information}), $\Psi$ for the regularizer,  $\lambda$ for the regularization strength, and  $\Omega$ for the search space. The fidelity term $\Phi(L,Q)$ enforces the influence of the initial labeling $L$, i.e., it decreases when $Q$ lies closer to $L$. The regularizer $\Psi(Q)$ favors geometrically smooth solutions, i.e., it enforces smoothness to all neighboring supervoxels ${v_{i}}$ and ${v_{j}}$, undergoing the same transformation $\mathbf{T}$. 
 $\Psi(Q)$is based on a Potts penalty function: 
\begin{equation} \label{eq:potts penalty}
\Psi_{Potts}(Q) = \begin{cases}
  1 & \text{ if  ${v_{i},v_{j}}$  consistent under $\mathbf{T}$}\\ 
  0 & \text{otherwise} 
  \end{cases} \enspace .
\end{equation}

\begin{figure*}
\centering
\includegraphics[scale=0.6,trim={0cm 5.5cm 8cm 0.7cm},clip]{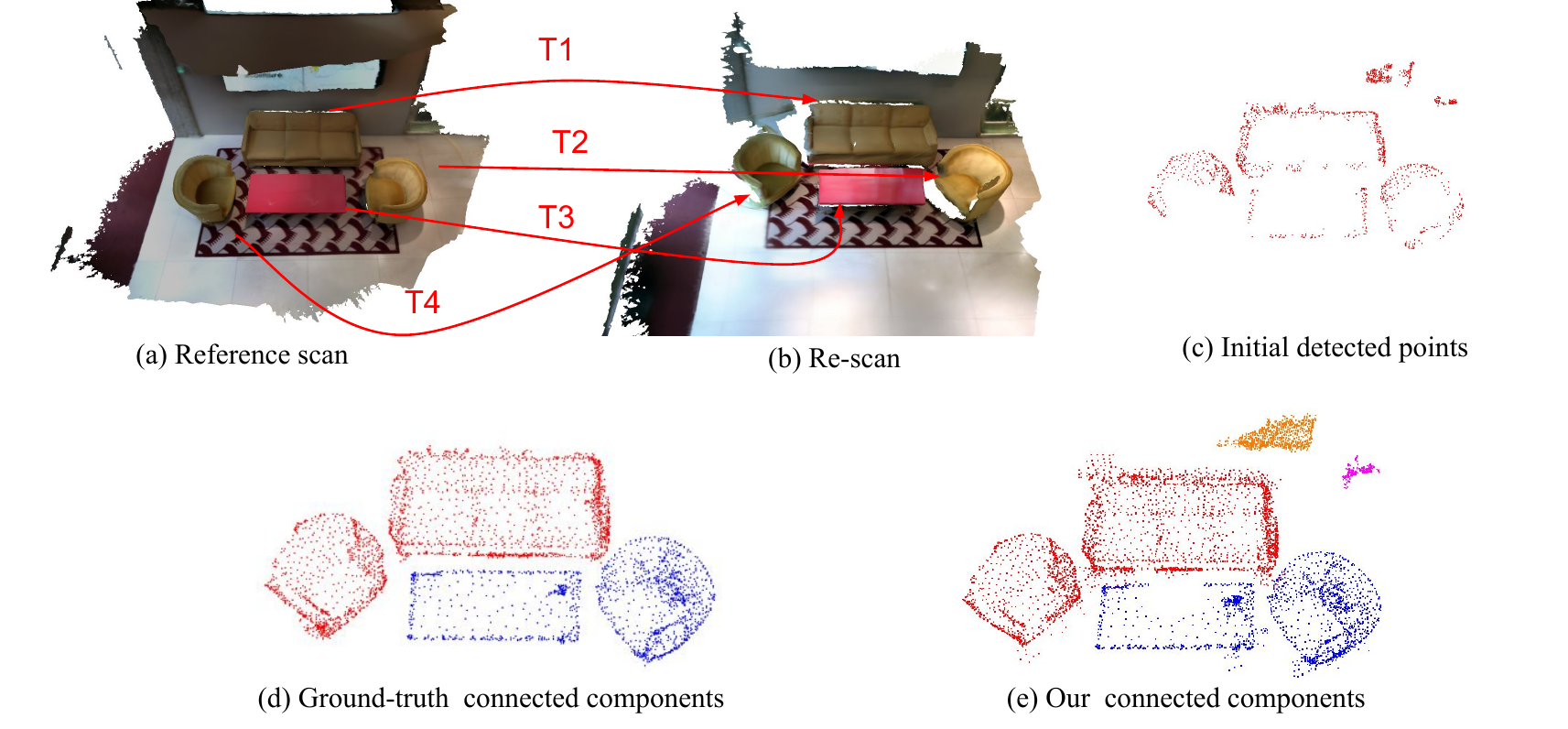} 
\caption{
Our approach: given two scans depicting a scene that has potentially changed, we discover all changes on an object-level.
We initially detect potentially changed scene parts by comparing depth maps. We then propagate changes and segment out changed regions based on the principle of geometric transformation consistency. (a) Reference scan (b) Rescan (c) Initially detected areas, with  false detections on the wall due to misalignments between the scans. (d) Ground truth connected components and (e) connected components detected by our approach}
\label{fig:overall}
\centering
\end{figure*}

It is important to note here that the energy function from Equation \ref{eq:optimized labeling} is conditioned on each computed transformation $\mathbf{T}$. Thus, we iterate through the top $k$ computed dominant transformations and we solve a series of graph cuts problems. Each iteration segments out the object undergoing the specific transformation $\mathbf{T}$. Objects added or removed, for which a transformation $\mathbf{T}$ is not established, are solely retrieved, based on the unary potentials of the optimization. The results of the iterative procedure, i.e., the set of points labeled as changing, are finally fused. A connected component analysis is finally applied to the fused results, to discover the final 3D objects. Connected component analysis is crucial to form the 3D added or removed objects that are not conditioned on a transformation, but also to overcome the problem of over-segmentation when slightly different transformations are computed for the same object. Optimization results are illustrated in Figures~\ref{fig:concept},~\ref{fig:overall} and~\ref{fig:results}.

\section{Experimental Evaluation} \label{Experimental}
\PAR{Datasets.} To assess the performance of the proposed approach, we have conducted experiments using the 3RScan dataset \cite{wald2019rio}. The dataset comprises individual rooms, capturing natural changing indoor environments. It provides, apart from the 3D meshes of the reference scans and the rescans, a series of RGB-D images captured by a Google Tango mobile phone and information concerning objects that have changed between the scenes, along with corresponding transformations. The experiments have been conducted on the validation subset of the dataset comprising 47 different reference scans and 110 rescans.  It is important to note that the 3RScan dataset was built initially for object instance relocalization tasks. Therefore, we had to generate the ground truth data for the changing objects based on the dataset's supplementary information. The code is publicly available at \href{URL}{https://github.com/katadam/ObjectsCanMove}, to enable the usage of this dataset for benchmarking indoor 3D change detection.

To the best of our knowledge, there is no other appropriatbe benchmark to assess 3D indoor change detection and 3D object discovery. Relevant works evaluate their methods on their own datasets, which are either not publicly available \cite{finman2013toward,langer2017fly,katsura2019spatial} or are very small and require manual labeling, as they do not provide appropriate annotations \cite{fehr2017tsdf,mason2012object,ambrus2016unsupervised}. Please refer to the SM for specific information on discarded datasets.

\PAR{Hyperparameter Tuning.} Ten randomly selected scans from the training split of 3RScan were used for parameter tuning, while the validation split was used for evaluation. The validation split covers many different scene types (i.e., offices, restaurants, living rooms, kitchens, etc.) to assess the generalization performance and robustness to challenging conditions and unseen environments. In our method, the main hyperparameters that need to be tuned are: the RANSAC inlier threshold $t$ for computing transformations $\mathbf{T}$, the number $k$ of transformations $\mathbf{T}$ to compute, and the weights for the graph optimization (as described in Section \ref{optimization}). The threshold $t$ can be set intuitively by the desired resolution of the transformations. The number $k$ of transformations should be set to the number of objects that change in a scene. Overestimating $k$ is not an issue as beyond the actual number of objects, RANSAC will be applied to outliers. Alternatively, one could also just stop once only a few matches are left or once the best model found by RANSAC only has a few inliers.

\begin{figure*}
\centering
\includegraphics[scale=0.48,trim={0.5cm 4.8cm 0cm 0cm},clip]{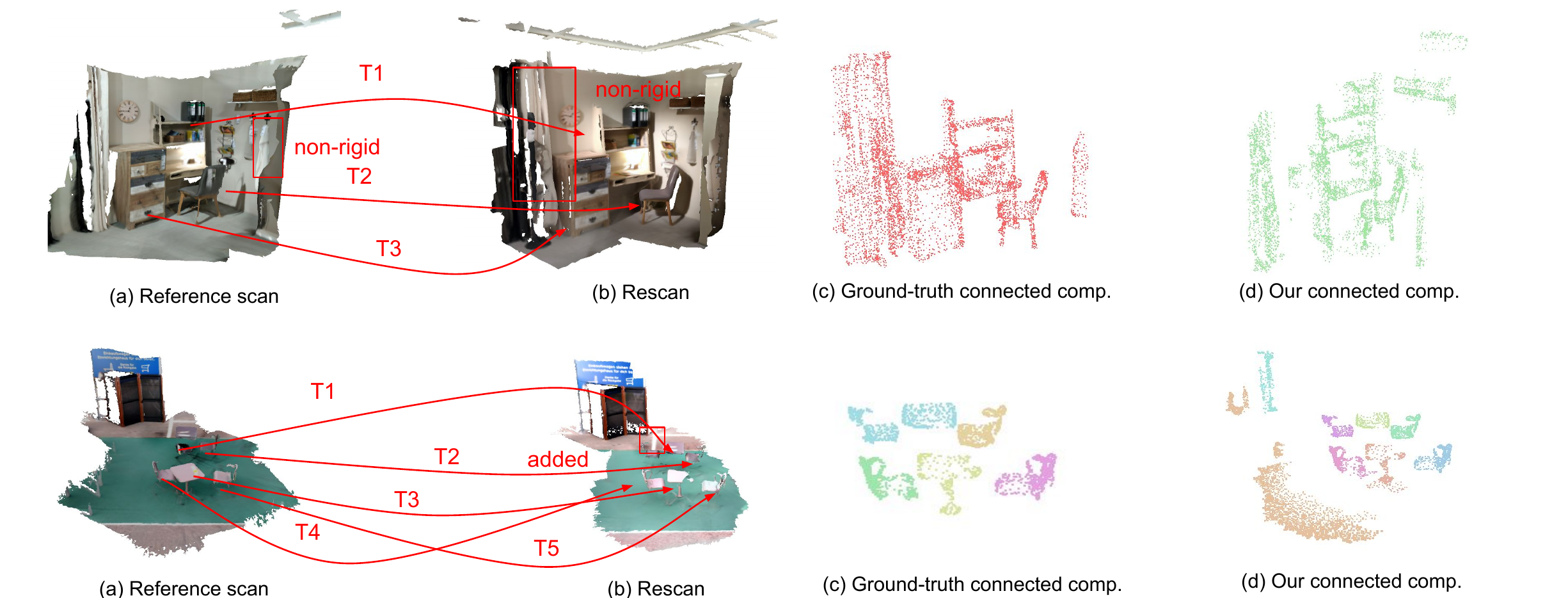} 
\caption{Qualitative evaluation of the proposed method. Given two scans (a reference scan (a) and a rescan (b)), we perform change detection on an object-level basis, to discover 3D objects. We visualise the final results after applying connected component analysis to the ground truth (c) and to our detected changes (d)}
\label{fig:results}
\centering
\end{figure*}


\PAR{Baseline Methods.} 
To compare the performance of our novel framework against a competitive set of other methods, we have searched for appropriate baselines. However,  we had to discard some works treating indoor change detection and unsupervised  object discovery, since they are not directly comparable to our method. More particularly, the input to our method are two scans, with changes between them but no recorded actual motion in front of the camera. 
As such, it is complementary to SLAM methodologies and technologies built upon SLAM systems \cite{finman2013toward,herbst2011toward,ambrus2016unsupervised,fehr2017tsdf}. Moreover, even though \cite{halber2019rescan,wald2019rio} deploy their methods on changing indoor scenes, they focus on instance segmentation and object instance re-localization, respectively. Thus, they cannot be evaluated against our change detection task. Approaches like \cite{langer2020robust,mason2012object} integrate semantics, contrary to our approach that discovers object-level changes without having a predefined notion of what an object is. The input of \cite{katsura2019spatial} is data from range sensors and a highly precise 3D map created by a 3D laser scanner, which is not the case for the 3RScan dataset. \cite{langer2017fly} aims only at discovering novel objects in the scene, while our approach retrieves all the changed objects. Towards that, we have a created a sub-task of discovering only added objects and compare against \cite{langer2017fly}. Results are available in the SM.  Finally, since we aim at change detection on an object-level towards unsupervised object discovery, we compared against an unsupervised 3D object discovery method \cite{karpathy2013object}. \cite{karpathy2013object} first discovers segments and then classifies them into objects and non-objects. However, the segments obtained via the authors' code after tuning parameter were not meaningful and we were not able to avoid a severe oversegmentation. Thus, we did not include the metrics in our experimental results. For visualizations  please refer to the SM.

Our approach is mainly inspired by the change detection approach from~\cite{taneja2011image,palazzolo2017change,palazzolo2018fast}. To the best of our knowledge, these are the most closely related baseline and one of our motivations to redefine this problem in a new framework by taking advantage of modern representations (i.e., supervoxels) and more recent graph optimization algorithms \cite{landrieu2017cut}. Similar to our work, \cite{palazzolo2018fast,palazzolo2017change,taneja2011image} are also focusing on unsupervised change detection. Taking all the above into consideration, we have decided to create two main baselines inspired by these works.

In \cite{taneja2011image}, change detection is based on inconsistency maps, formed by subtracting pairs of images taken at different points in time. The newly acquired image is warped into the old one, using the known 3D scene geometry and the known poses of both images with respect to the scene. Assuming similar illumination conditions, the two images should be identical if no change in the geometry has happened. In turn, changes in scene geometry will lead to inconsistent projections from one image into the other. Change detection is then optimized via a graph cut on the voxelized representation. The inconsistency maps are used to calculate the unary term of the graph, while the binary term accounts for color smoothing. Similar to the first step of \cite{taneja2011image}, \cite{palazzolo2018fast,palazzolo2017change} are discovering changes by formulating inconsistency maps. These works augment the number of inconsistency maps to achieve better results without any further optimization.

Since two 3D models are available in our case, we use the initial change detection step from Section~\ref{initial_dect} to create the inconsistency maps for the two baselines inspired by~\cite{taneja2011image,palazzolo2017change,palazzolo2018fast}. We go one step further and resort to depth images instead of RGB images to ensure robustness to illumination conditions. This initial change detection step (i.e., our method before optimization) serves as the 1st baseline, namely \textbf{Papazzolo et al.}, as it is  equivalent to the work presented in \cite{palazzolo2018fast,palazzolo2017change}. In these works, estimation of 3D change detection results from back-projecting inconsistencies from multiple 2D maps.

A 2nd baseline (\textbf {Taneja et al.}) is formed, following~\cite{taneja2011image}, where the initial change detection is optimized ensuring color consistency on a voxelized representation of the scene via a graph cut optimization (solved by max-flow algorithm \cite{boykov2004experimental}). The binary term of the graph is computed as described in Equation \ref{eq:5}:
\begin{equation} \label{eq:5}
\psi_{ij}(l_i,l_j) = [l_{i} \neq l_{j}]\cdot\gamma/(\sum_{{I}_{t}}{ ||{v_{t}^{i}-v_{t}^{j}}||^2} +1),
\end{equation}
where $||{v_{t}^{i}-v_{t}^{j}}||^2$ accounts for the L2-norm between RGB values of voxels $v_{t}^{i}$ and $v_{t}^{j}$ and $\gamma$ is a regularization factor. Comparing against this baseline shows the impact of using geometric consistency for propagating change, which is the main technical contribution of this work.

\PAR{Ablation Study.}Three more baselines are formed in the form of an ablation study, for a better insight into the proposed method. Ablation baselines are reporting intermediate results of our framework. They also calculate the metrics when the method has access to more information, in order to test its robustness with respect to different parameters. Removing the optimization part of our method and relying only on initial change detection is equivalent to \cite{palazzolo2018fast} and thus reported in Table~\ref{Tab1a}. The first ablation baseline (\textbf{ground truth transforms.}) ensures geometric consistency using the ground truth transformations provided by the dataset instead of our computed ones. This gives an upper bound to the performance we can achieve and helps measure the impact of estimated transformation's accuracy on the overall system's efficiency.

\begin{figure*} 
\centering
\includegraphics[scale=0.6,trim={1cm 6cm 4.5cm 4.5cm},clip]{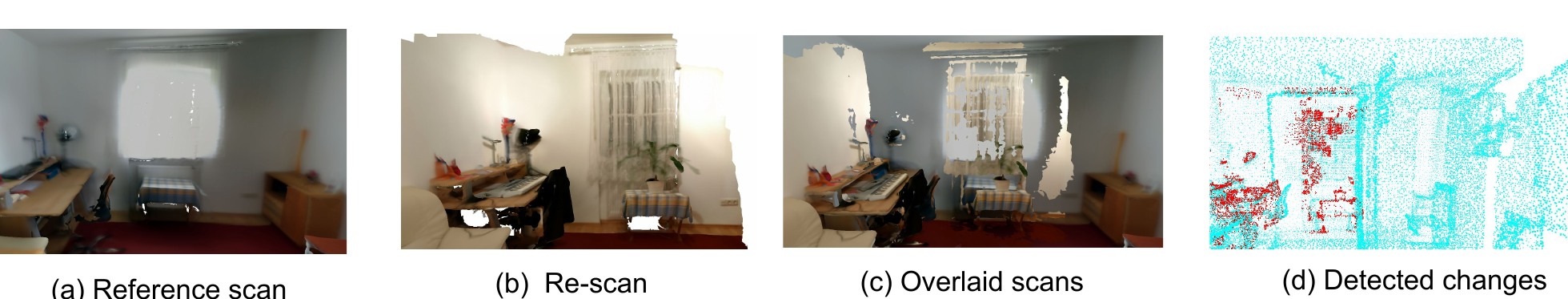}
\caption{A non-rigid change (curtain) is not recorded in the ground truth. The curtain is different between the reference (a) and rescan (b), as shown when the two scans are overlaid in (c). The detected changes are shown with red colour in (d), overlaid on the reference scan in blue}
\label{curtain}
\centering
\end{figure*}

As the 2nd ablation baseline (\textbf{{RANSAC inliers}}), closely related to \cite{steinhauser2008motion}, we present the metrics of the non-static points used to form the matches and compute the rigid transformations $\mathbf{T}$. This is equivalent to only the second step of our method (Section \ref{Transformations}), without the initial change detection and the graph optimization. Ideally, each set of inliers consistent with each RANSAC execution would form the corresponding object moved under this transformation. 
 
Finally, as the 3rd ablation baseline (\textbf{{Mask-RCNN}}), we add a semantic component to the formulated algorithm, as we would like to get an idea of how well our approach performs with respect to a supervised method. Thus, we replace our novel geometric consistency-based term with a term based on the instance labels of Mask-RCNN~\cite{he2017mask}, propagating the change to all regions sharing the same semantic label. Mask-RCNN  is a powerful method for 2D object detection. We deploy the 2D object detector, trained on the COCO dataset \cite{lin2014microsoft}, on the RGB images of each rescan.

\begin{table}[t!]
\centering
\small{
\caption{Mean IoU and mean recall for the proposed method and the published baselines}
\begin{tabular}{l|l|l} 
Method                                      & IoU(\%) & Recall(\%)  \\ \hline  \hline
Palazzolo et al.\cite{palazzolo2018fast} / Ours bf optim.            &     54.23\%  & 31.48\%                          \\ \hline
Taneja et al. \cite{taneja2011image}                        &    48.10\%     &      44.50\%        \\ \hline
Our method                        &    68.40\%    &76.05\%  
{\label{Tab1a}}\end{tabular}
}
\end{table}

\begin{table}[t!]
\centering
\caption{Mean IoU and mean recall for the proposed method and the ablation study's baselines}
\small{
\begin{tabular}{l|l|l} 
Method                                      & IoU(\%) & Recall(\%)  \\ \hline  \hline
Our method                        &    68.40\%    &76.05\%  \\\hline
Ground truth transforms.                            &   72.40\%      &      93.89\%        \\ \hline
Mask-RCNN                            &    52.96\%    &       89.22\%     \\ \hline
RANSAC inliers                           &    10.82\%    &       29.50\%      
{\label{Tab1b}}\end{tabular}
}
\end{table}


\PAR{Experimental Results.} 
In addition to the qualitative results presented in Figures~\ref{fig:concept},~\ref{fig:overall},~\ref{fig:results}, we rigorously evaluate our method by using metrics that capture the success of 3D change detection and  3D object discovery. Since we are aiming at object discovery through change detection on an object-level basis, we should first evaluate the efficiency of our change detection results. Thus, we calculate the metric of recall, on a voxel basis. Recall aims to calculate how many of the ground truth changed voxels have been correctly retrieved. 

Moving on to 3D object discovery, a 3D connected components analysis is applied to the change detection results. To assess the efficiency of the proposed method as an object discovery pipeline, we deploy the metric of Interestion Over Union (IoU) per discovered object, as it encapsulates both the metrics of precision and recall. To calculate this metric, the connected component analysis is also applied to the ground truth changes. For our scenario, this analysis was performed on a voxel grid of 10~cm, which could sometimes merge objects that lie together into a single component. This does not affect our metrics since the same connected component analysis is applied both to the ground truth and our solution. However, a smaller step size would lead to a more refined and detailed object discovery. The parameter can be tuned based on the size of the objects we want to discover. We consider an object  as successfully discovered when the metric of IoU is more than $20\%$.  The metrics are calculated at a voxel-level since we are interested in measuring how two objects (volumes) intersect.

Tables \ref{Tab1a} and \ref{Tab1b} show the mean recall over all the scans and the mean IoU of discovered objects. After close examination, it is clear that our method outperforms the most competitive baseline based on \cite{taneja2011image} by roughly 30\% in terms of recall. It also improves the mean IoU by almost 20\%. This shows that not only supervoxels constitute a more efficient representation compared to single voxels, when it comes to graph optimization, but also that the novel geometric transformation consistency is much more successful for propagating change, compared to  photoconsistency. Moreover, evaluation metrics before and after graph optimization, demonstrate the importance of the optimization, as it improves the mean IoU by 14.17\% and the mean recall by 44.57\%. As explained above, the method presented in \cite{palazzolo2017change,palazzolo2018fast} is equivalent to the first step of our method, thus showing improved performance of our presented framework over all published baselines. Integrating a voxel graph cut optimization, propagating change to color-consistent regions \cite{taneja2011image}, leads to better recall rates, but lower IoU, as change is in some cases overpropagated, resulting in low precision, an thus failure of discovering the objects, in terms of IoU. 

Concerning the ablation, as denoted by the results of the MASK-RCNN baseline, adding a semantic component is not improving the overall performance. The MASK-RCNN baseline  is capable of achieving a mean IoU of 52.96\% and a mean recall of 89.22\%. This can be attributed to noisy RGB-D images, leading to inaccurate segmentations. Indeed, background patches are falsely detected as foreground objects. Thus, change is propagated into a large percentage of the scene's background, leading to a higher recall rate, compared with a relatively low precision, and thus IoU. The solution using the ground truth transformation (baseline ground truth transforms.) is in close proximity with our method in terms of recall. Even a coarser estimation of the rigid transformation of the scene is capable of achieving close to the best possible results. However, there is still space for improvement, regarding the computation of transformations. The mean IoU of 72.40\% in this baseline, is explained due to initial false detections, caused by occlusions and misalignments between scans. The experimental results indicate that the two scans need to be correctly registered to avoid false initial detections. False initial detections are merged with correctly estimated regions, reducing the IoU score. Finally, it is worth mentioning that using only non-static parts discovered by RANSAC iterations leads to results worse than our solution before graph cuts optimization. This explicitly demonstrates that the straightforward approach of  feature matching and computing sets of motion-consistent points is insufficient.

\begin{figure*}[t!]
\centering
\includegraphics[scale=0.58,trim={2.2cm 4cm 2.2cm 5cm},clip]{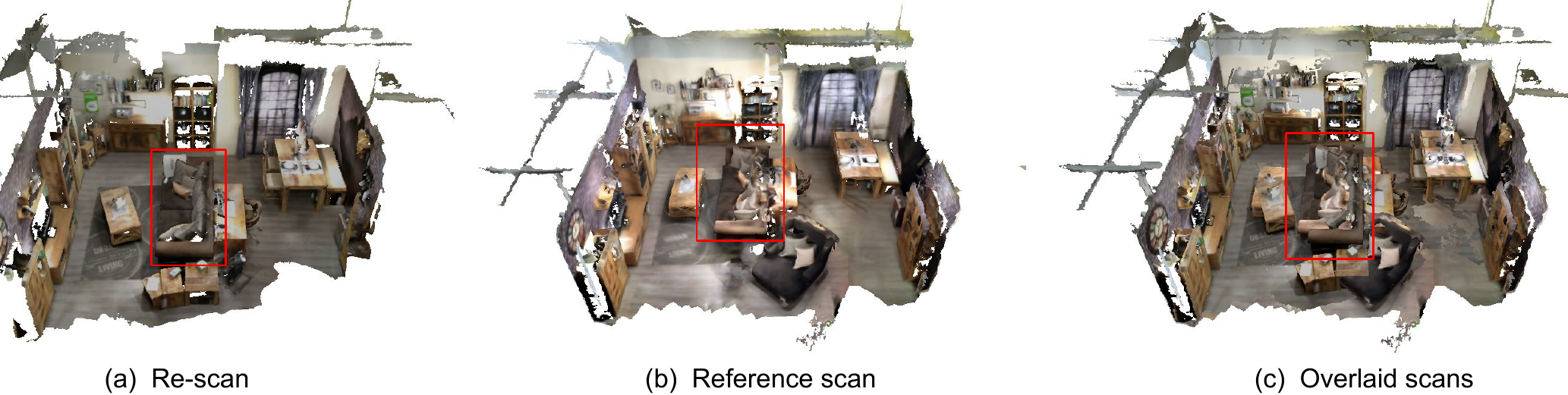} 
\caption{A moved couch between the reference scan (a) and the rescan (b) that is not part of the ground truth annotations. Overlaid scans in (c)}
\label{fig22}
\centering
\end{figure*}

Finally,  3RScan is a dataset built towards assessing object instance relocalization  and not exhaustive change detection. Thus, our method uncovers changes between the scans not recorded in the ground truth. Such cases would affect the evaluation metrics, and we wanted to check their extent. We randomly selected a subset of 10 rescans and visually inspected them. In 60.00\% of cases, we discovered an unrecorded change (see, for example, Figure \ref{fig22}). The proposed approach has correctly detected 66,67\% of these cases. Moreover, an example of a non-rigid and not recorded change is depicted in Figure \ref{curtain}. 

\PAR{Limitations.} By definition of discovering objects via change detection, we will miss objects that do not undergo a substantial enough change. Using a stricter threshold for distinguishing between inliers and outliers in the RANSAC scheme could help recover even small motions. Moreover, depth map  subtraction could lead to false initial change detection when the two scans are not entirely aligned. A typical example is illustrated in the second row of Figure~\ref{fig:results}. Parts of the floor are labeled as changing, forming a 3D object due to the scan's misalignment.

\section{Conclusion}
The presented method achieves state-of-the-art performance on the object discovery task, via change detection on an object-level basis, for the 3RScan dataset against a competitive set of baselines. The method shows the surprising effectiveness of using scene change for high-recall object discovery and of using motion constraints to achieve precise detections. The very general assumption that objects are connected and move in a coherent way is used to propagate initial detections. Importantly, these geometric cues can be discovered directly from unannotated data, so they do not introduce strong priors or any memorization of what objects are.

\PAR{Acknowledgment.} This research was supported by projects EU RDF IMPACT No. CZ.02.1.01/0.0/0.0/15\_003/0000468, EU H2020 ARtwin No. 856994 and the EU
Horizon 2020 project RICAIP (grant agreement No 857306).

\begin{appendix}

\title{Appendix}
\author{}
\institute{}
\maketitle

\noindent This is the appendix for our paper ``Objects can move: 3D Change Detection by Geometric Transformation Consistency".  In Section~\ref{Quanative}, we discuss more quantitative results for the task of 3D object discovery, and provide a more thorough investigation of the calculated transformations. 
In Section \ref{qualitative}, more visual results of our method  and the baselines 
are shown. We showcase corner cases where our output is inconsistent with the dataset's ground-truth annotation. 
We discuss in Section~\ref{sec:datasets} why 3RScan \cite{wald2019rio} is the most appropriate dataset for evaluating 3D change detection and 3D object discovery, and we also evaluate our results on the sub-task of discovering added objects 
(Section \ref{added}).

\section{Quantitative Results} \label{Quanative}

\PAR{Accuracy of the Computed Transformations.} As explained in the main paper, the computation of the transformations induced by moving objects constitutes an essential component of the proposed method. We extract DGCNN features~\cite{wang2019dynamic} and establish correspondences on the whole scene based on nearest neighbor search. Transformations are then computed using an iterative RANSAC procedure~\cite{bolles1981ransac}. We further evaluate the accuracy of the rigid transformations with respect to the ground-truth ones. Results shown in Table \ref{transformationMetrics}, are evaluated in terms of recall, capturing the percentage of correctly calculated transformations. We also provide the Mean Translation (MTE) and Mean Rotation (MRE) Error, for all the correctly retrieved 3D rigid transforms. As in \cite{wald2019rio}, we consider a transformation as successfully calculated, if the alignment errors for the translation $t_{\Delta}$ and rotation $R_{\Delta}$ are lower than $10cm, 10^{\circ}$ and $20cm, 20^{\circ}$ respectively. 

\begin{table*}[]
\caption{Transformation evaluation via the percentage of poses within given error bounds on the position and orientation error (Recall), the Median Translation Error (MTE) (in meters), and the Median Rotation Error (MRE) (in degrees). MRE and MTE are provided for the correctly retrieved transformations, i.e., when the alignment errors for the translation $t_{\Delta}$ and rotation $R_{\Delta}$ are lower than $10cm, 10^{\circ}$ and $20cm, 20^{\circ}$ respectively} 
\label{transformationMetrics}
\small{
\begin{tabular}{P{2.2cm}|P{1.5cm}|P{1.5cm}|P{1.5cm}|P{1.5cm}|P{1.5cm}|P{1.5cm}|}

                Method &Recall (\textless{}0.10m, 10$^{\circ}$) & MRE(deg) & MTE(m) &Recall (\textless{}0.20m, 20$^{\circ}$)& MRE(deg) & MTE(m) \\ \hline \hline
FPFH~\cite{rusu2009fast}         & 2.61                                               & 7.25                          & 0.0645                      & 8.36                                             & 10.57                        & 0.0776                      \\ \hline
SHOT~\cite{tombari2010unique}           & 6.79                                               & 5.35                          & 0.0268                      & 12.27                                            & 8.18                         & 0.0393                      \\ \hline
3D-match dynamic ~\cite{zeng20173dmatch}   & 5.48                                               & 5.81                          & 0.0542                      & 13.05                                            & 7.30                          & 0.0708                      \\ \hline
RIO static~\cite{wald2019rio}       & 9.92                                               & 4.33                          & 0.0425                      & 17.75                                            & 6.39                          & 0.0545                      \\ \hline
RIO dynamic~\cite{wald2019rio}    & 15.14                                              & 4.75                          & 0.0437                      & 23.76                                            & 6.08                          & 0.0547                      \\ \hline
Our method             & 3.58                                               & 3.00                          & 0.0799                      & 18.21                                            & 4.25  & 0.1381                      \\ 

\end{tabular}
}
\end{table*}

We compare our approach with handcrafted and learned descriptors for establishing correspondences. The baseline methods follow the object instance re-localization protocol of the 3RScan dataset~\cite{wald2019rio}: having access to an instance segmentation of the reference scan, they only need to find correspondences for 3D parts belonging to each instance. In contrast, our method does not use this supervisory signal and performs full matching between the scenes.  Thus, the included baselines have access to more information compared to our approach. As expected, the baselines estimate more precise transformations. However, our results show that our approach is still competitive with such strong baselines.

\PAR{Ablation of the $k$ Transformations  Used in the Optimization.} One of the tunable hyperparameters of our method is the number of top $k$ transformations $\mathbf{T}$ used to propagate changes during the optimization. Indeed, sometimes wrong transformations with few inliers that are caused by imperfect correspondences are established. We thus ablate the top $k$ transformations with the most inliers to propagate the change to neighboring regions. Table \ref{Tab2} shows the results for the proposed method when $k = 5, 10, 15$ transformations are used. As expected, using a larger $k$ increases the mean recall (i.e., the percentage of correctly retrieved objects) and decreases the mean IoU, as in some cases, the change leads to oversegmentation. However, there is no substantial difference in the overall performance of the proposed method correlated with $k$. This validates the robustness of our approach.

\begin{table}[]
\centering
\caption{\label{tab:table-name}Mean IoU and mean recall for the proposed method using different number of computed transformations $k$ to propagate geometrical consistency} {\label{Tab2}}
\small{
\begin{tabular}{P{2.5cm}|P{2.5cm}|P{2.5cm}|} 
Number of trans.                                    & IoU(\%) & Recall(\%)  \\ \hline  \hline
5            &     68.40\%  & 76.05\%                          \\ \hline
10                    &    64.89\%     &     77.43\%        \\ \hline
15                       &    65.50\%    &79.09\%  \\
\end{tabular}
}
\end{table}

\begin{table}[ht!]
\centering{
\small{
\caption{Components of each presented method}
\begin{tabular}{P{5.5cm}|P{1.5cm}|P{1.5cm}|P{1.5cm}} 
Method                                      & Init. Detect. & Comp. Transf. & Optim. 
\\ \hline  \hline
Palazzolo et al. \cite{palazzolo2018fast}  /Ours bf optim.                         &    \checkmark   &  &   \\ \hline 
Taneja et al.  \cite{taneja2011image}                     &     \checkmark   &  &  \checkmark\\ \hline
Our method            &     \checkmark   &  \checkmark &  \checkmark             
{\label{baselines}}\end{tabular}
}
}
\end{table}

\PAR{Accuracy and Completeness of the 3D Discovered Objects.} Towards assessing 3D object discovery, we also deploy the metrics of accuracy and completeness (on the point level). Per object accuracy refers to the percentage of correctly predicted 3D points out of all the points forming our discovered object. On the other hand, completeness captures how many of the ground-truth object's points were correctly retrieved by our solution. Tables \ref{Tab2a} and \ref{Tab2b} show the mean accuracy and mean completeness for all objects. After close examination, it is clear that our proposed method balances the most between the two metrics when compared with the two published 
baselines (Palazzolo et al., Taneja et al.). Table \ref{baselines} summarizes the different components of each published baselines. Concerning the ablation baseline, having access to the ground-truth transformations would slightly improve the overall performance. On the other hand, assigning semantics masks to many parts of the scene (Mask-RCNN), i.e., labeling most of the scene as foreground objects, results in  a high completeness rate. However, it also leads to poor accuracy as change is wrongly propagated to static regions with the same label, starting from wrong initial change detections.

\begin{table}[t!]
\centering
\caption{Mean accuracy and mean completeness for the proposed method and the published baselines}
\small{
\begin{tabular}{P{6cm}|P{2cm}|P{2cm}|} 
Method                                      & Acc.(\%) & Compl.(\%)  \\ \hline  \hline
Palazzolo et al. \cite{palazzolo2018fast}  /Ours bf optim.                     &    67.76 \%  & 33.39\%                           \\ \hline
Taneja et al.  \cite{taneja2011image}                         &    65.97\%     &     30.53\%        \\ \hline
Our method                        &    54.60\% &59.20\%  
\label{Tab2a}
\end{tabular}
}
\end{table}

\begin{table}[t!]
\centering
\small{
\caption{Mean accuracy and mean completeness for the proposed method and the ablation study's baselines}
\begin{tabular}{P{6cm}|P{2cm}|P{2cm}|}
Method                                      & Acc.(\%) & Compl.(\%)  \\ \hline  \hline
Our method                        &    54.60\% &59.20\%  \\\hline
Ground-truth transforms.                            &   48.79\%      &     74.54\%        \\ \hline
Mask-RCNN                            &    37.43\%    &      71.39\%     \\ \hline
RANSAC inliers                           &    52.47\%    &      14.73\%    
{\label{Tab2b}}
\end{tabular}
} 
\end{table}

\section{Qualitative Results} \label{qualitative}
In the following, initial detections of changing regions are depicted along with the graph cut optimization results and ground-truth annotations. Corner cases are also discussed.

\begin{figure*}[]
\centering
\includegraphics[scale=0.45,trim={0cm 2cm 3cm 1.7cm},clip]{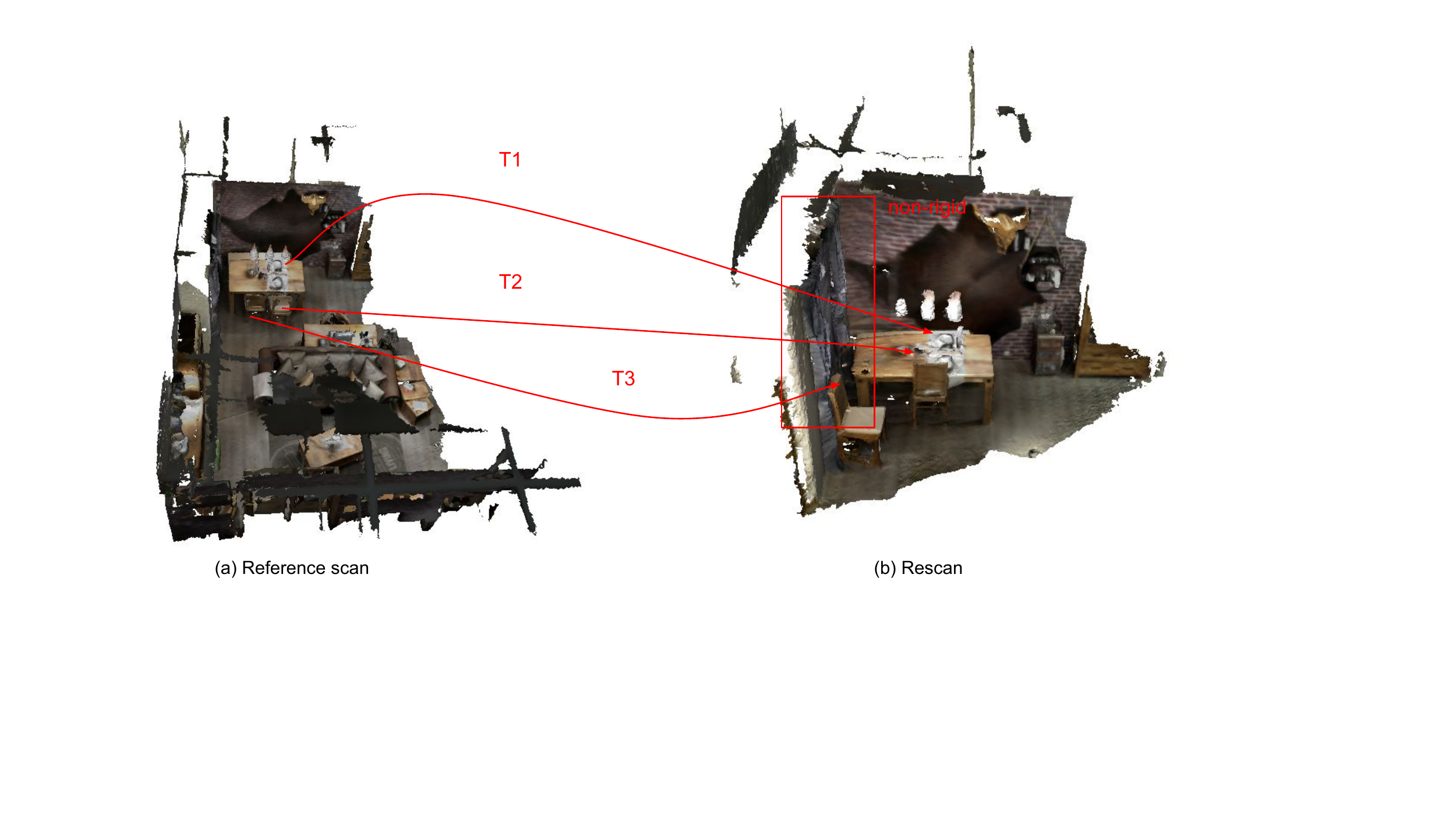} 
\includegraphics[scale=0.45,trim={0cm 5cm 0cm 1.7cm},clip]{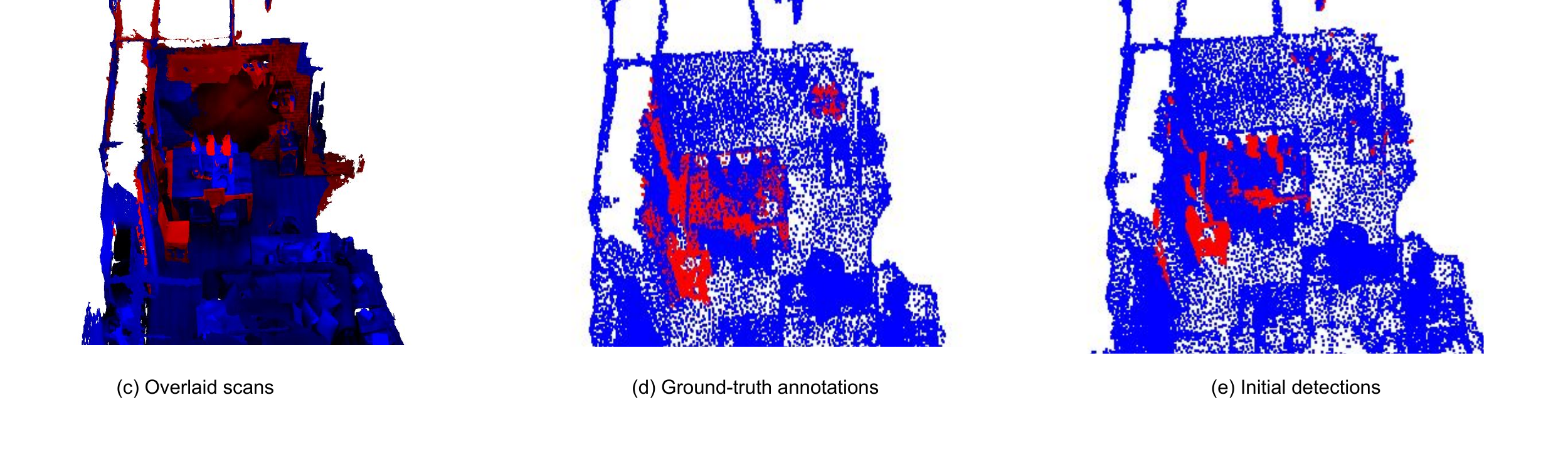} 
\caption{Reference scans (a) and re-scans (b) of multiple scenes with highlighted occurred changes (T1, T2, \, etc. refer to the ground-truth transformations between moving objects). Overlaid meshes of reference scan and rescan in red and blue (c). Ground-truth changed regions of the point clouds (in red) overlaid on the
reference scan (in blue) (d). Initial change detection results of the point clouds (in red) overlaid on the reference
scan (in blue) (e)}
\label{initialA1}
\centering
\end{figure*}

\begin{figure*}
\centering
\includegraphics[scale=0.45,trim={0cm 3cm 0cm 0cm},clip]{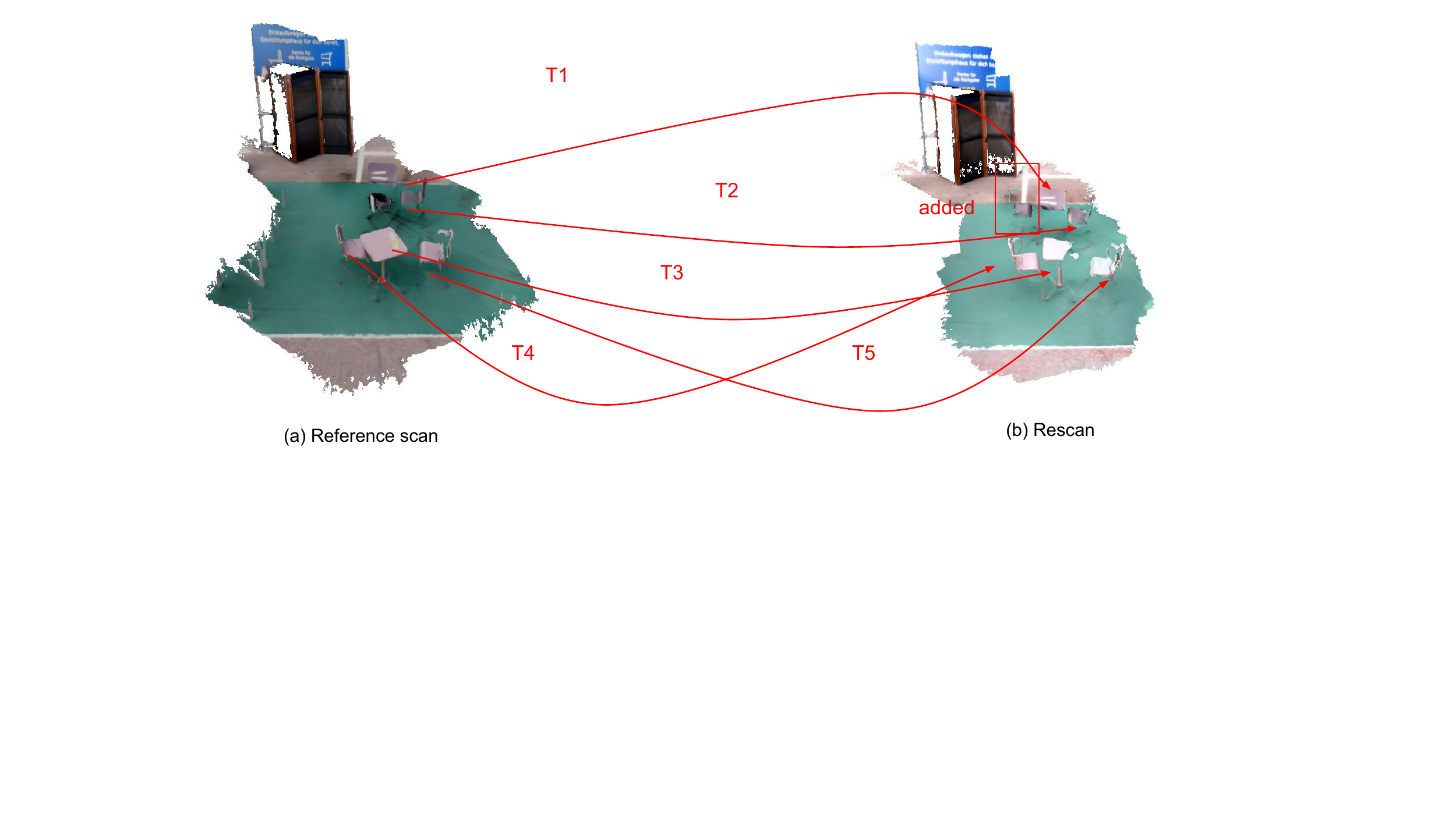} 
\includegraphics[scale=0.45,trim={0cm 6cm 1cm 0.6cm},clip]{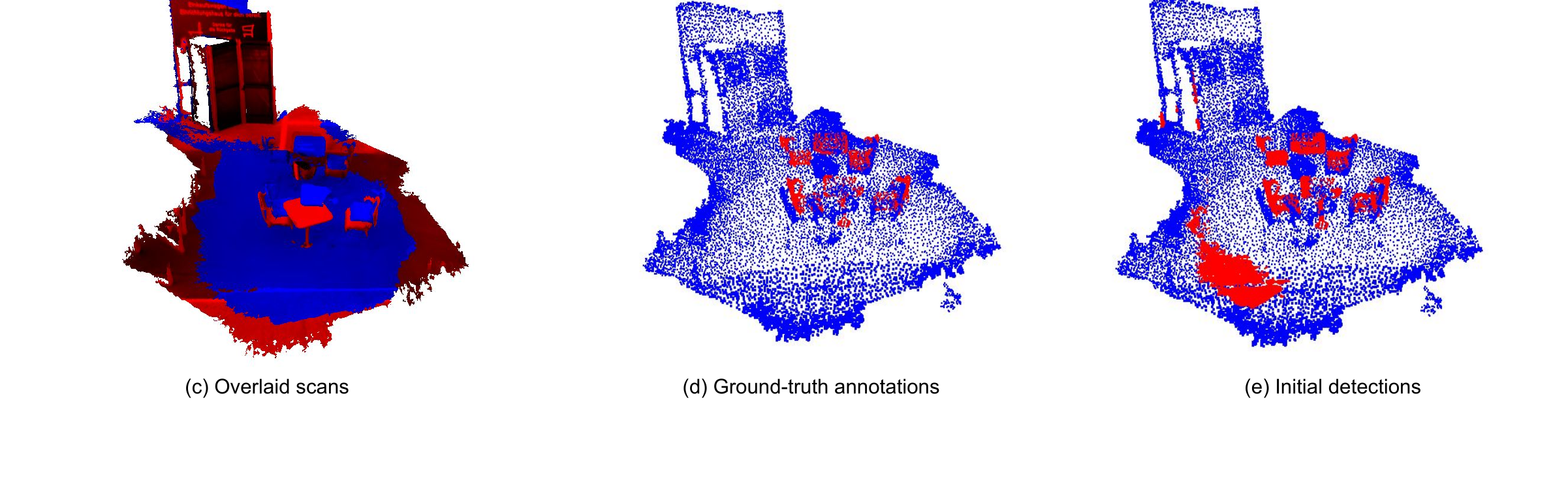} 
\caption{Reference scans (a) and re-scans (b) of multiple scenes with highlighted occurred changes (T1, T2, etc. refer to the ground-truth transformations between moved objects). Overlaid meshes of reference scan and rescan in red and blue (c). Ground-truth changed regions of the point clouds (in red) overlaid on the
reference scan (in blue) (d). Initial change detection results of the point clouds (in red)  overlaid on the reference
scan (in blue) (e)}
\label{initialB1}
\centering
\end{figure*}

\begin{figure*}[htb!]
\centering
\includegraphics[scale=0.45,trim={0cm 4cm 0cm 0cm},clip]{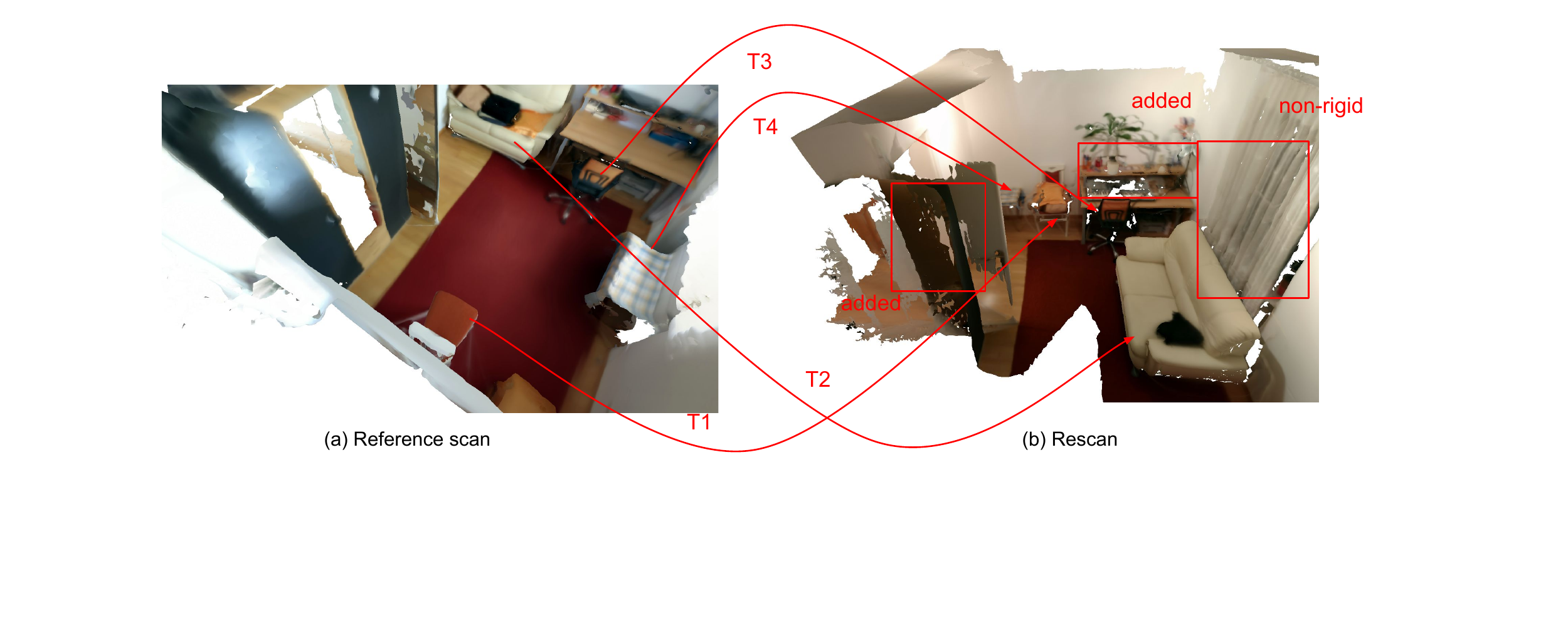} 
\includegraphics[scale=0.45,trim={0cm 5cm 1cm 0.6cm},clip]{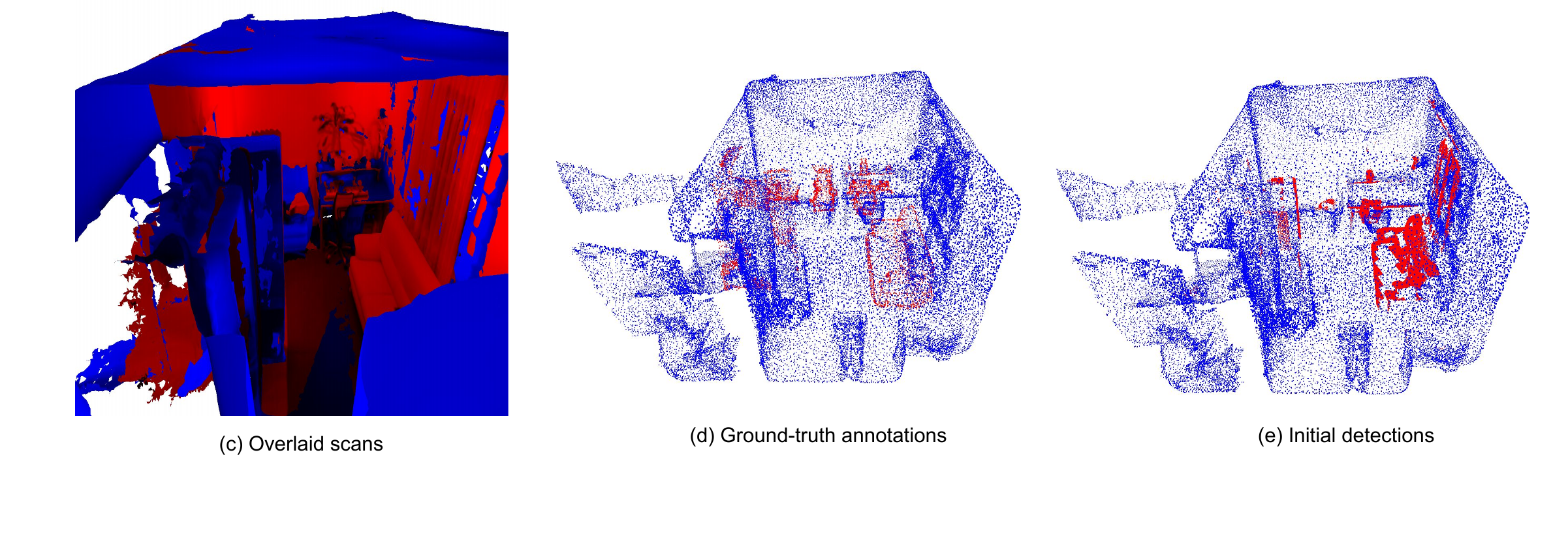} 
\caption{Reference scans (a) and re-scans (b) of multiple scenes with highlighted occurred changes (T1, T2, etc.. refer to the ground-truth transformations between moved objects). Overlaid meshes of reference scan and rescan in red and blue (c). Ground-truth changed regions of the point clouds (in red) overlaid on the
reference scan (in blue) (d). Initial change detection results of the point clouds (in red) overlaid on the reference
scan (in blue) (e)}
\label{initialB2}
\centering
\end{figure*}

\PAR{Initial Detection Results.} Results of the initially discovered changing regions are depicted in Figures \ref{initialA1}, \ref{initialB1}, and \ref{initialB2}. 
It is important to note here that in most  cases, the rescans constitute only partial observations of the reference scans; thus, the marked changing regions refer only to the parts visible in the rescan. After close inspection of Figures  \ref{initialA1}, \ref{initialB1}, and \ref{initialB2}, it becomes clear that in most  cases, our initial detection stage efficiently retrieves the changing regions. Wrong detections are primarily attributed to slight misalignments between the reference scan and the rescan. 

\begin{figure*}[htb!]
\centering
\includegraphics[scale=0.45,trim={0cm 5cm 3cm 0cm},clip]{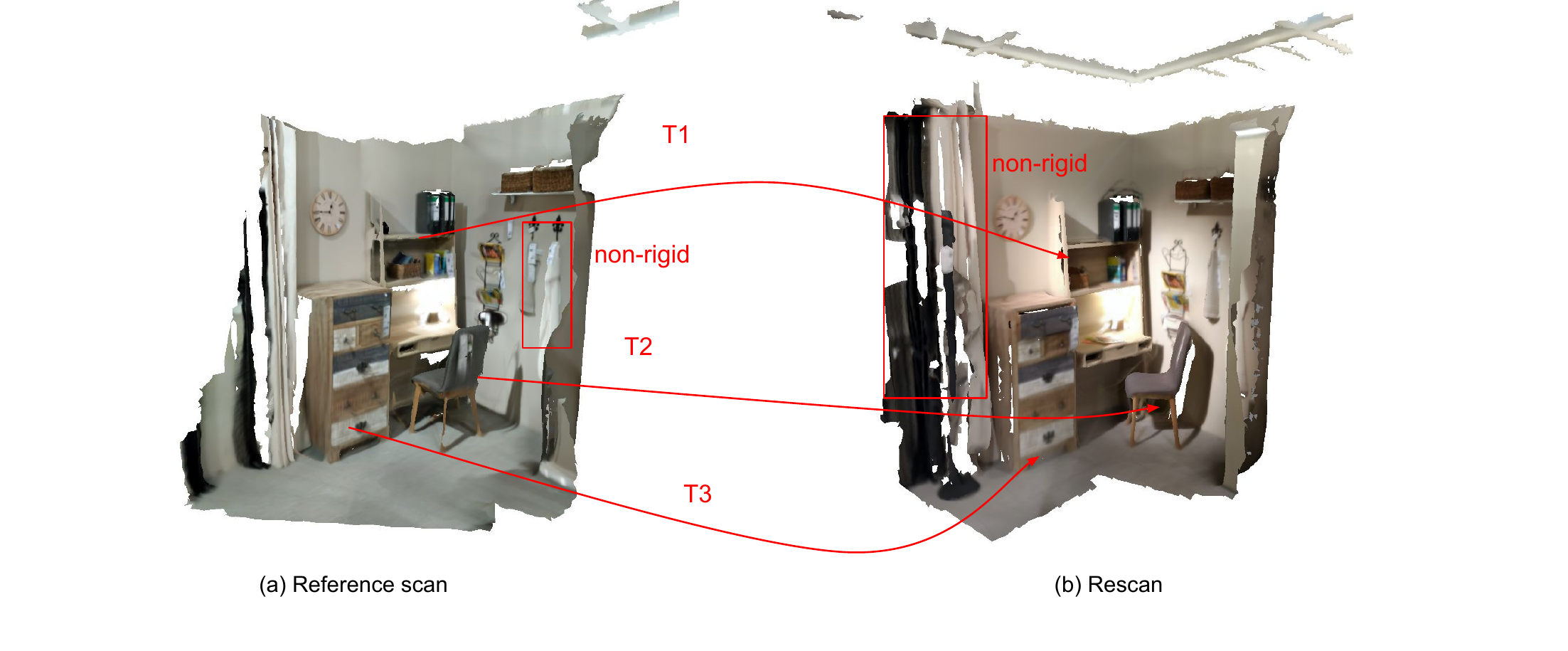} 
\includegraphics[scale=0.45,trim={0cm 4cm 0cm 0cm},clip]{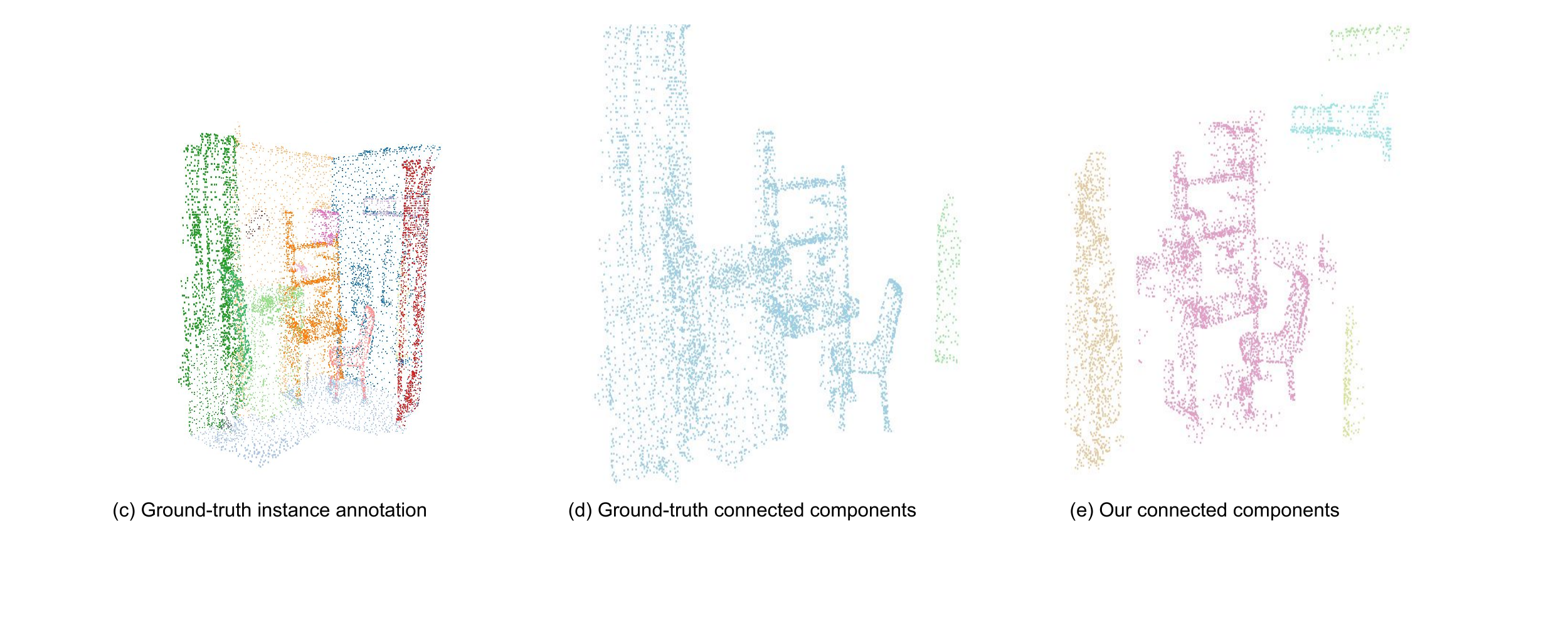} 
\caption{The meshes of the reference scan (a) and the rescan (b). Ground-truth instance segmentation of the point cloud of the rescan (c). Ground-truth connected components of the point cloud (d), compared with the connected components of our solution in (e)}
\label{componentsA2}
\centering
\end{figure*}
\begin{figure*}[htb!]
\centering
\includegraphics[scale=0.45,trim={0cm 4cm 0cm 0cm},clip]{30.pdf} 
\includegraphics[scale=0.45,trim={1cm 6cm 0cm 1cm},clip]{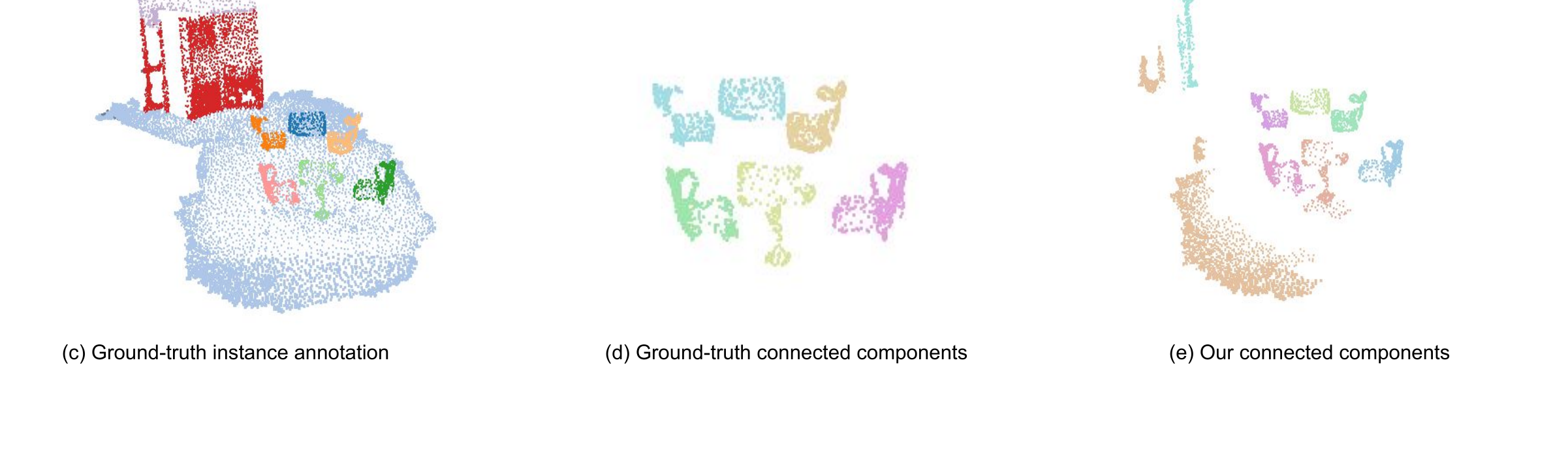} 
\caption{The meshes of the reference scan (a) and the rescan (b). Ground-truth instance segmentation of the point cloud of the rescan (c). Ground-truth connected components of the point cloud (d), compared with the connected components of our solution in (e)}
\label{componentsB1}
\centering
\end{figure*}
\begin{figure*}[htb!]
\centering
\includegraphics[scale=0.45,trim={0cm 5cm 0cm 0cm},clip]{40.pdf} 
\includegraphics[scale=0.45,trim={0.2cm 6cm 0cm 0cm},clip]{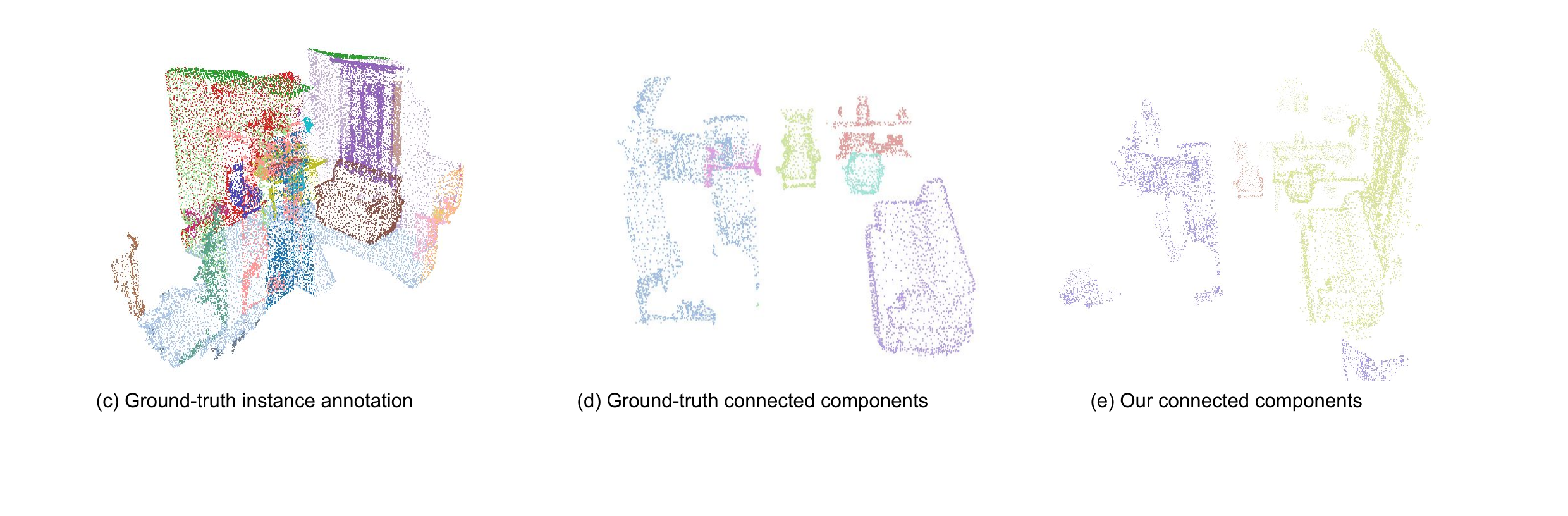}  
\caption{The meshes of the reference scan (a) and the rescan (b). Ground-truth instance segmentation of the point cloud of the rescan (c). Ground-truth connected components of the point cloud (d), compared with the connected components of our solution in (e)}
\label{componentsB2}
\centering
\end{figure*}

\PAR{Results after graph cut Optimization.} After graph cut optimization, a 3D connected component analysis is applied both to the ground-truth annotations and to the results of our approach. 
This step aims to turn our detected changing regions into the final discovered objects. Figures \ref{componentsA2}, \ref{componentsB1}, and \ref{componentsB2} present the final output of the proposed method. 



\PAR{Corner Cases.} We also present cases where our results conflict with the annotations provided by the dataset. This could be partially attributed to unrecorded changes in the ground-truth.  3RScan~\cite{wald2019rio} is a dataset built towards assessing object instance re-localization and thus does not exhaustively record every change.   
 Typical examples are shown in Figures \ref{componentsB2}  and \ref{incosistency}, where the non-rigid change of the curtain is not recorded. However, there are also cases where rigid transformations are not included in the ground-truth annotations (cf. the refrigerator in Figure \ref{incosistency}). 
 In all the above-mentioned cases, the proposed method successfully detected these changes. 

The main limitation of our method is handling misalignments between scans. Indeed, differences in voxel occupancy and depth values occur when the reference scan and the rescan are not correctly registered. Thus, regions are falsely labeled as changing during our method's initial detection step. A typical example is Figure \ref{falseDetection}, where parts of the floor are wrongly retrieved as changing due to the misalignment. Post-processing steps such as matching against a 3D object database, can eliminate these false detections.

\begin{figure*}[]
\centering
\includegraphics[scale=0.45,trim={0cm 0cm 0cm 0cm},clip]{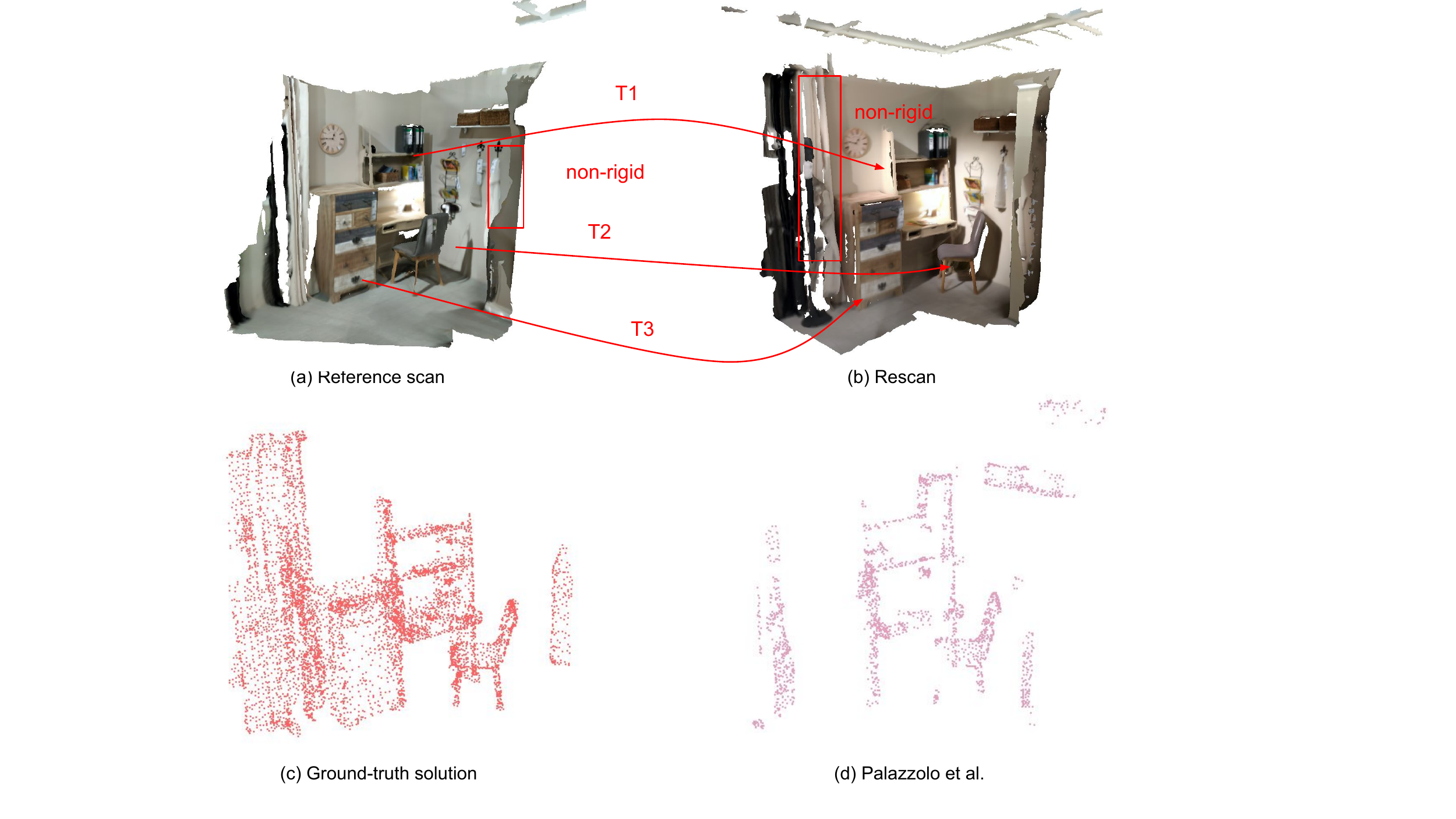} 
\includegraphics[scale=0.45,trim={0cm 0cm 0cm 0cm},clip]{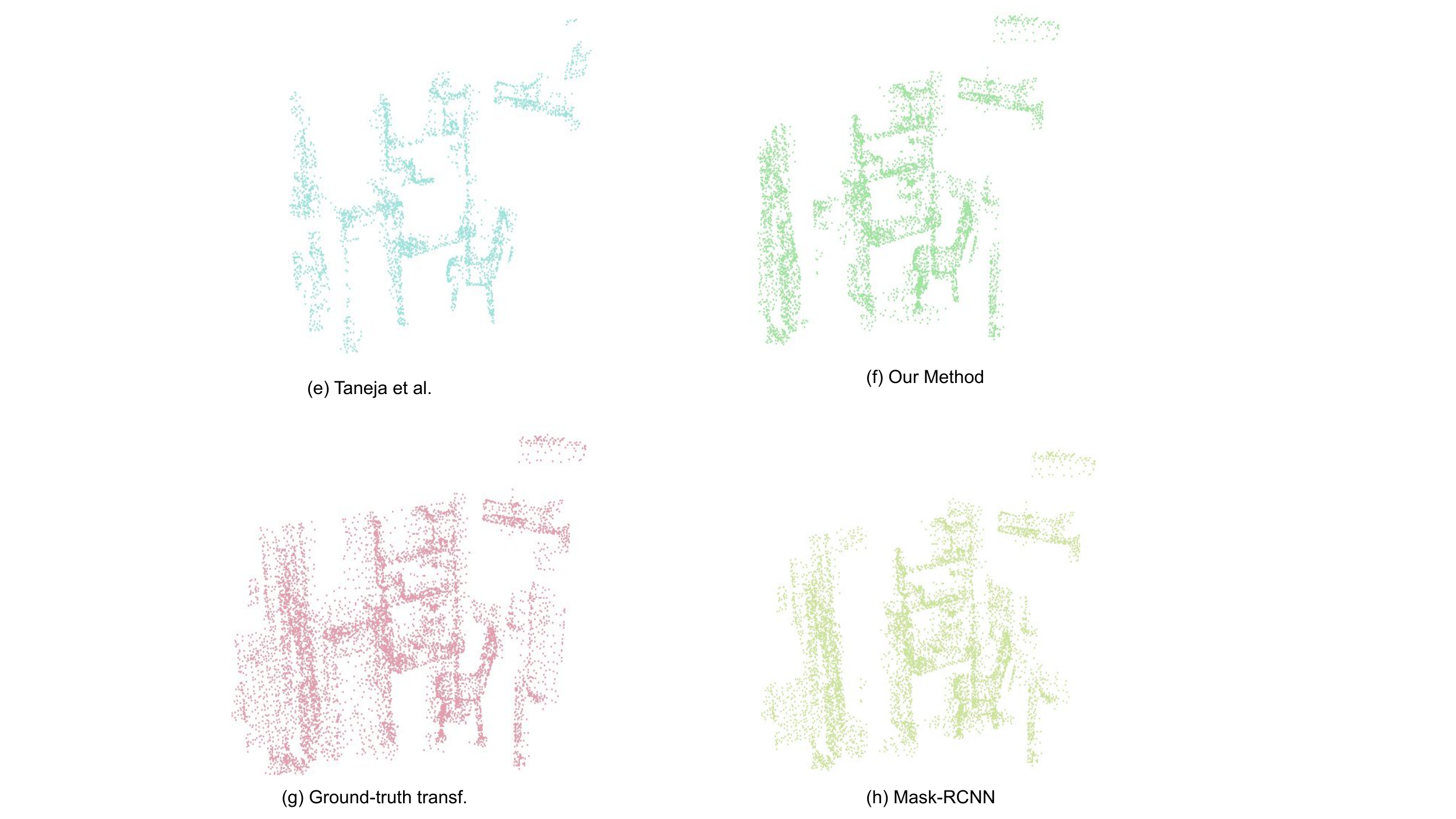} 

\caption{The meshes of the reference scan (a) and the rescan (b). Ground-truth solution (c). Results of the published baselines in (d) and (e). Results  of the proposed method in (f). Results of the ablation study's baselines in (g) and (h)}
\label{baselineComparisonB}
\centering
\end{figure*}

\begin{figure*}[h!]
\centering
\includegraphics[scale=0.45,trim={0cm 0cm 0cm 0cm},clip]{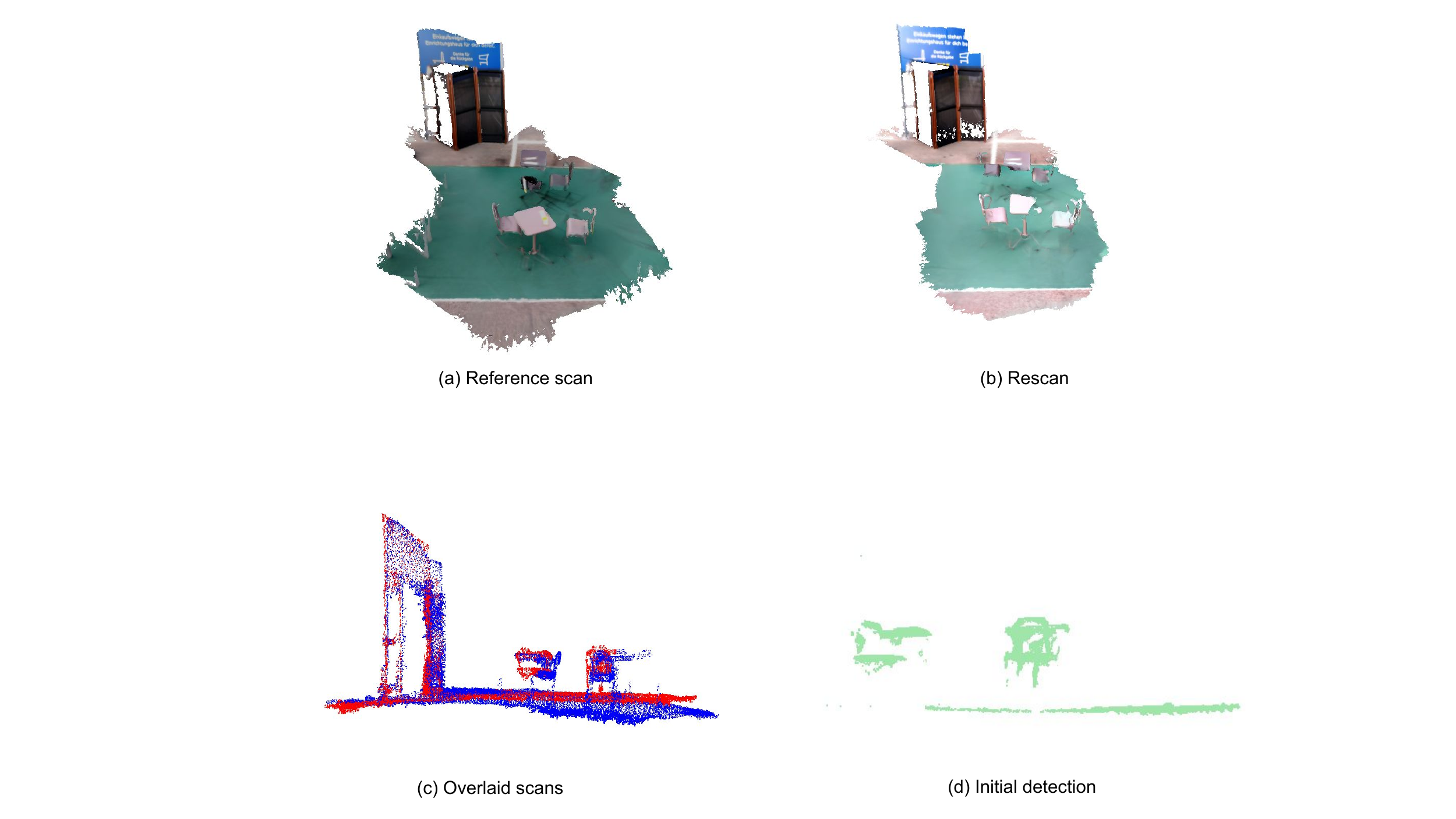} 
\caption{Slight misalignment between reference scan (a) and rescan (b). Overlaid reference scan and rescan (depicted in blue and red respectively) in (c). Initial detection results in (d)}
\label{falseDetection}
\centering
\end{figure*}

\begin{figure}[ht!]
\centering
\includegraphics[scale=0.45,trim={0cm 7cm 0cm 1cm},clip]{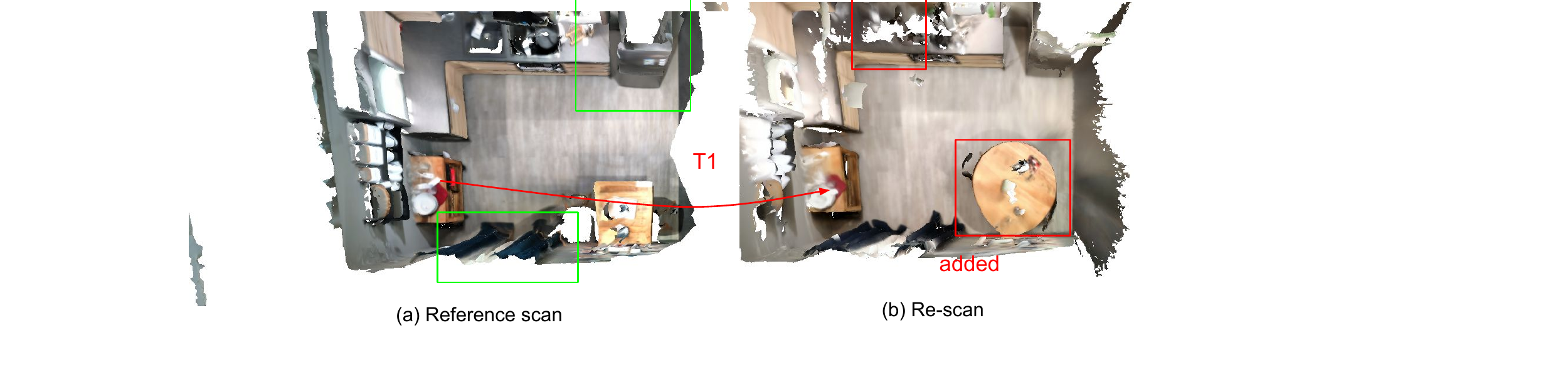} 
\includegraphics[scale=0.45,trim={0cm 8cm 0cm 0cm},clip]{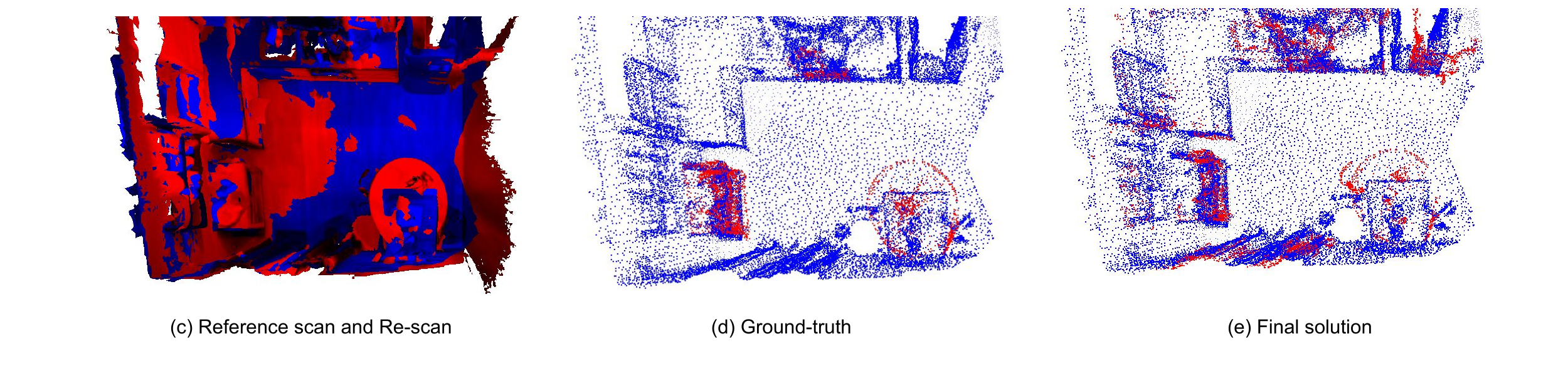} 
\caption{Changes not recorded in the ground-truth. 
The mesh of the reference scan is depicted in (a), and the green bounding boxes underline unrecorded changes. The mesh of the rescan in (b). Recorded changes in the ground-truth are underlined in red color. The meshes of the two scans overlaid in red and blue color, respectively (c). Ground-truth annotations (in red) overlaid on the point cloud of the reference scan in blue (d), and our graph cut optimization results (in red) overlaid on the point cloud of the reference scan in blue (e). The rigid change of the refrigerator and the non-rigid change of the curtain that were not recorded in the ground-truth are successfully detected by our method}
\label{incosistency}
\centering
\end{figure}

\PAR{Baseline Comparisons.} As stated in the Section 4 of the main paper, we compare our method against two published baselines, Palazzolo et al. \cite{palazzolo2018fast} and Taneja et al.~\cite{taneja2011image}. 
We also compare against three baselines in the form of an ablation study (Ground-truth transforms, Mask-RCNN, RANSAC Inliers). Qualitative results for the two published baselines and the ablation baselines (Ground-truth transforms and Mask-RCNN) are depicted in Figure \ref{baselineComparisonB}.

The main paper explains that Palazzolo et al.~\cite{palazzolo2018fast} is equivalent to our method before the optimization step, i.e., it predicts change through depth comparison. The depicted visual results prove the need for a more sophisticated solution rather than simply relying on inconsistencies between 2D projections. On the other hand, Taneja et al.~\cite{taneja2011image} performs a graph cut optimization, where the binary term ensures photoconsistency. This constraint seems to perform well on objects with homogeneous texture, such as the chair in Figure \ref{baselineComparisonB}, but fails when the same object has multiple textures (cf. the cabinet in Figure \ref{baselineComparisonB}). 

Moving on to the ablation study, the baseline using the ground-truth transformations provided by the dataset seems to work better than the presented method. This is expected since our method firstly computes transformations and then discovers objects. Thus, for the proposed method, the uncertainty of the calculated transformations is propagated to the results of the graph optimization step. Finally, it is evident that when Mask-RCNN is used, a lot of background regions are labelled as changing, since they were wrongly extracted as foreground in the RGB-D images (i.e., they were incorrectly assigned a semantic mask). Moreover, it completely fails in regions that were not detected as a foreground object (cf. cabinet in Figure \ref{baselineComparisonB} ), due to the lack of relevant training data for these concepts. In contrast, our method does not rely on any predefined notion of what an object should look like. 

\section{Datasets}
\label{sec:datasets}

The 3RScan dataset~\cite{wald2019rio} is a dataset built towards assessing object instance re-localization under rigid transformations. Thus, the dataset provides information about the transformations induced by moving objects, along with an instance segmentation of each scene. It also provides information about non-rigid changes in some instances. 
Hence, we generate the ground-truth annotations by (i) extracting all the information about rigid and non-rigid movements as provided by the dataset, and (ii) by comparing the instance segmentation of the reference scan and rescan to discover objects that have been added or removed. 

As stated above, even though 3RScan is not directly designed for 3D change detection/ 3D object discovery, ground-truth information can be generated without manual labeling. It is also a very large and diverse dataset, and thus, to the best of our knowledge, the most appropriate dataset for our needs. Table \ref{Tab1datasets} summarizes the characteristics of the adopted dataset, against other applicable datasets that were not suitable for our scenario.

\begin{table}[t!]
\centering
\caption{Applicable datasets for change detection/ object discovery} {\label{Tab1datasets}}
\begin{tabular}{P{2.7cm}|P{1.5cm}|P{1.5cm}|P{1.5cm}|P{1.7cm}|P{1.5cm}|}
Method & Scans & Rescans & Instance Seg. & Annotation & Available  \\ \hline  \hline
Finman et al.\cite{finman2013toward}             &   2   &   67  &  ?   &     ?          & \textemdash   \\ \hline
Langer et al.\cite{langer2017fly}            &     1       &      4   &   \textemdash   &  \textemdash    &  \textemdash\\ \hline
Katsura et al.\cite{katsura2019spatial}                         &    2     & 10+?     &  ? & ?  & \textemdash    \\ \hline
Herbst et al.\cite{herbst2011toward} &     4    &    24  & \checkmark  &  \checkmark     & \textemdash\tablefootnote{The provided URL is not valid}\\ \hline
Mason et al.\cite{mason2012object} &  1       &  67    &  \textemdash  &   \textemdash    & \checkmark\tablefootnote{Data available upon request}\\ \hline
Ambrus et al.\cite{ambrus2016unsupervised} &   1      &   88   & \textemdash  & \textemdash\tablefootnote{Inconsistent and incomplete annotation}   & \checkmark   \\ \hline
Fehr et al.\cite{fehr2017tsdf} & 3 & 23 & \textemdash  & \textemdash   & \checkmark   \\ \hline
Wald et al.\cite{wald2019rio} - 3RScan &   478      &   1482   & \checkmark &  \checkmark  &\checkmark\tablefootnote{With our provided code}  \\ \hline
Halber et al.\cite{halber2019rescan} &   13      &    45  & \checkmark & \textemdash  &\checkmark    \\ \hline
Langer et al.\cite{langer2020robust} &    5     &   31   & \textemdash  & \checkmark  & \checkmark
\end{tabular}
\end{table}

As shown in Table \ref{Tab1datasets}, there is no  widely available dataset appropriate for evaluating 3D indoor change detection and 3D object discovery. Each work evaluates its efficiency on tailor-made datasets, some of which \cite{finman2013toward,langer2017fly,katsura2019spatial,herbst2011toward} are not publicly available. Concerning the public datasets, \cite{mason2012object} provides a relative large amount of rescans. However, only a single environment is considered. Moreover, its objects are not annotated so it cannot be used for quantitative evaluation. Similarly, \cite{ambrus2016unsupervised} consists of data captured by a robot in an office setting. The diversity of the provided scenes is thus limited. The annotation is mostly inconsistent, since some selected objects are annotated as new  while other objects that are physically new in a scene are not. In a similar vein to our used dataset \cite{wald2019rio}, \cite{fehr2017tsdf} uses a hand-held Google Tango device to capture three rooms (reference scans) and 23 rescans. Taking into account that the dataset is much smaller and less diverse and its complete lack of annotations, we have decided not to use it. The authors of \cite{langer2020robust} have created their own dataset, for evaluating added small objects (from the YCB dataset \cite{gadde2017efficient}) in the scene. \cite{langer2017fly} provides ground-truth annotation only for novel objects and not for moved ones, which makes the quantitative evaluation of our task hard as not all the cases we are interested in can be directly tested. Finally, \cite{halber2019rescan} aims at tracking instance segmentation across temporal changes. Thus, a ground-truth instance annotation is provided for every rescan, but no annotations concerning moving/static objects. 

\section{Novel Objects} \label{added}
Novel objects are also of broad interest to multiple robotic applications. Since existing works in the research community \cite{langer2017fly,langer2020robust} focus on novel added objects, we have decided to create a subtask of discovering all added objects in the scene. To this end, we have prepossessed 3RScan to create a new ground truth including only the novel objects, and we also decided to compare our algorithm against \cite{langer2017fly}. Originally, \cite{langer2017fly} detects only small objects on the floor. We modified this work to discover objects regardless of their size and position for a fair comparison. Table~\ref{novel} shows the results in terms of IoU and recall at the voxel level, to capture the volumetric overlap between ground truth and predictions. After close inspection of Tab.~\ref{novel}, it is clear that our method outperforms the most competitive baseline in the sub-task of discovering added objects. 

\vspace{-12pt}
\begin{table}[ht!]
\centering
\small{
\caption{Metrics of the proposed method and \cite{langer2017fly} on novel objects of the 3RScan dataset}
\begin{tabular}{P{2cm}|P{2cm}|P{2cm}} 
Method                                      & IoU(\%) & Recall(\%)  \\ \hline  \hline
\cite{langer2017fly}        &     73.85\%  & 64.25\%                          \\ \hline
Our method                        &    75.08\%    &76.70\%  
{\label{novel}}\end{tabular}
}\end{table}

\newpage

\newpage

\vfill
\break
\end{appendix}

\clearpage
%

\bibliographystyle{splncs04}
\bibliography{egbib}

\end{document}